%% file: ms.tex
\documentclass[lettersize, journal, twoside]{IEEEtran}

\IEEEoverridecommandlockouts                              
\makeatletter

\let\proof\@undefined
\let\endproof\@undefined

\let\NAT@parse\undefined
\makeatother

\input{style/definitions}
\title{
Robust Task Scheduling for Heterogeneous Robot Teams under Capability Uncertainty
}

\author{Bo Fu, William Smith, Denise Rizzo, Matthew Castanier, Maani Ghaffari, and Kira Barton

\thanks{Manuscript received 10 December 2021; revised 17 June 2022; accepted 22 September 2022.
This work was supported by the Automotive Research Center, a U.S. Army center of excellence for modeling and simulation of ground vehicles.
Distribution Statement A: Approved for public release; distribution unlimited (OPSEC 5970).}
\thanks{Bo Fu, Maani Ghaffari, and Kira Barton are with the Department of Robotics,  University of Michigan, Ann Arbor, MI 48109, USA (e-mail: bofu@umich.edu; maanigj@umich.edu; bartonkl@umich.edu)}
\thanks{William Smith, Denise Rizzo, and Matthew Castanier are with the US Army DEVCOM Ground Vehicle Systems Center, Warren, MI 48397, USA (e-mail: william.c.smith1019.civ@army.mil; denise.m.rizzo2.civ@army.mil; matthew.p.castanier.civ@army.mil)}%
}

\begin{document}

\maketitle

\input{section/abstract}

\input{section/introduction}

\input{section/related_work}
\input{section/model}

\input{section/optimization_algorithm}
\input{section/flow_cover}

\input{section/experiment_result}

\input{section/experiment_medical}

\input{section/conclusion}

\input{section/appendix.tex}

\bibliographystyle{IEEEtran}
\bibliography{style/ieee-abrv,style/strings-abrv,ms}

\clearpage
\input{section/biography}


\end{document}

%% file: style/definitions.tex
\usepackage{graphicx}
\usepackage[font={small}]{caption}
\usepackage[english]{babel}
\usepackage{amsmath,amssymb,enumerate}
\usepackage{mathtools}
\usepackage[mathscr]{euscript}
\usepackage{mathrsfs}
\usepackage{times}
\usepackage[usenames,dvipsnames,svgnames,table, xcdraw]{xcolor} 
\usepackage{amsfonts}
\usepackage{url}
\usepackage{subcaption}
\usepackage{booktabs}
\usepackage{multirow}
\usepackage[ruled,vlined,linesnumbered]{algorithm2e}
\SetKw{Continue}{continue}
\SetKw{Break}{break}
\let\labelindent\relax
\usepackage{enumitem}
\usepackage{cite} 
\usepackage{balance}
\usepackage[colorlinks=true,pdfpagemode=UseNone,citecolor=black,linkcolor=red,urlcolor=BrickRed]{hyperref}
\usepackage[normalem]{ulem}

\newif\ifshowcomment

\newcommand{\modelrisk}{{\ifshowcomment\color{Green}\fi CTAS}}
\newcommand{\modeldet}{{\ifshowcomment\color{Green}\fi CTAS-D}}
\newcommand{\modelint}{{\ifshowcomment\color{Green}\fi CTAS-O}}

\newcommand{\bigO}[1]{\Ocal(#1)} 

\newcommand{\Ocal}{\mathcal{O}}

\newcommand{\transpose}{\mathsf{T}}

\newcommand{\node}{N}

%% file: section/abstract.tex
\begin{abstract}

This paper develops a stochastic programming framework for multi-agent systems where task decomposition, assignment, and scheduling problems are simultaneously optimized.
The framework can be applied to heterogeneous mobile robot teams with distributed sub-tasks. Examples include pandemic robotic service coordination, explore and rescue, and delivery systems with heterogeneous vehicles.
Due to their inherent flexibility and robustness, multi-agent systems are applied in a growing range of real-world problems that involve heterogeneous tasks and uncertain information. Most previous works assume one fixed way to decompose a task into roles that can later be assigned to the agents. This assumption is not valid for a complex task where the roles can vary and multiple decomposition structures exist. Meanwhile, it is unclear how uncertainties in task requirements and agent capabilities can be systematically quantified and optimized under a multi-agent system setting.
A representation for complex tasks is proposed: agent capabilities are represented as a vector of random distributions, and task requirements are verified by a generalizable binary function. The conditional value at risk (CVaR) is chosen as a metric in the objective function to generate robust plans. An efficient algorithm is described to solve the model, and the whole framework is evaluated in two different practical test cases: capture-the-flag and robotic service coordination during a pandemic (e.g., COVID-19). Results demonstrate that the framework is generalizable, scalable up to 140 agents and 40 tasks for the example test cases, and provides low-cost plans that ensure a high probability of success.
\end{abstract}


%% file: section/introduction.tex
\section{Introduction}
\IEEEPARstart{T}{echnological} advances in sensing and control have enabled robotic applications in an ever-growing scope. On the other hand, the growing complexity and requirements of the applications soon exhaust the capability of a single robot: limited by design rules and actuator/sensor power, even the most competent robot is not able to handle all real-world tasks alone. This trend fosters the recent proliferation of multi-agent system applications in agriculture \cite{tokekar2016sensor}, warehouse management \cite{wurman2008coordinating}, construction \cite{werfel2014designing}, defense \cite{shishika2020cooperative}, exploration \cite{cai2021non, quann2017energy, quann2018ground, quann2019chance, quann2020power}, and surveillance \cite{schlotfeldt2018resilient, yu2019coverage, sung2020distributed}.
The advantages of replacing a large omnipotent robot with a team of smaller and less powerful robots include the robustness to agent failures, resilience of team configuration, and lower maintenance costs (a large robot with the same task capabilities is usually harder to design and costlier due to the system complexity) \cite{tan2013research}.

\begin{figure}[t]
    \centering
    \includegraphics[width=0.9\linewidth]{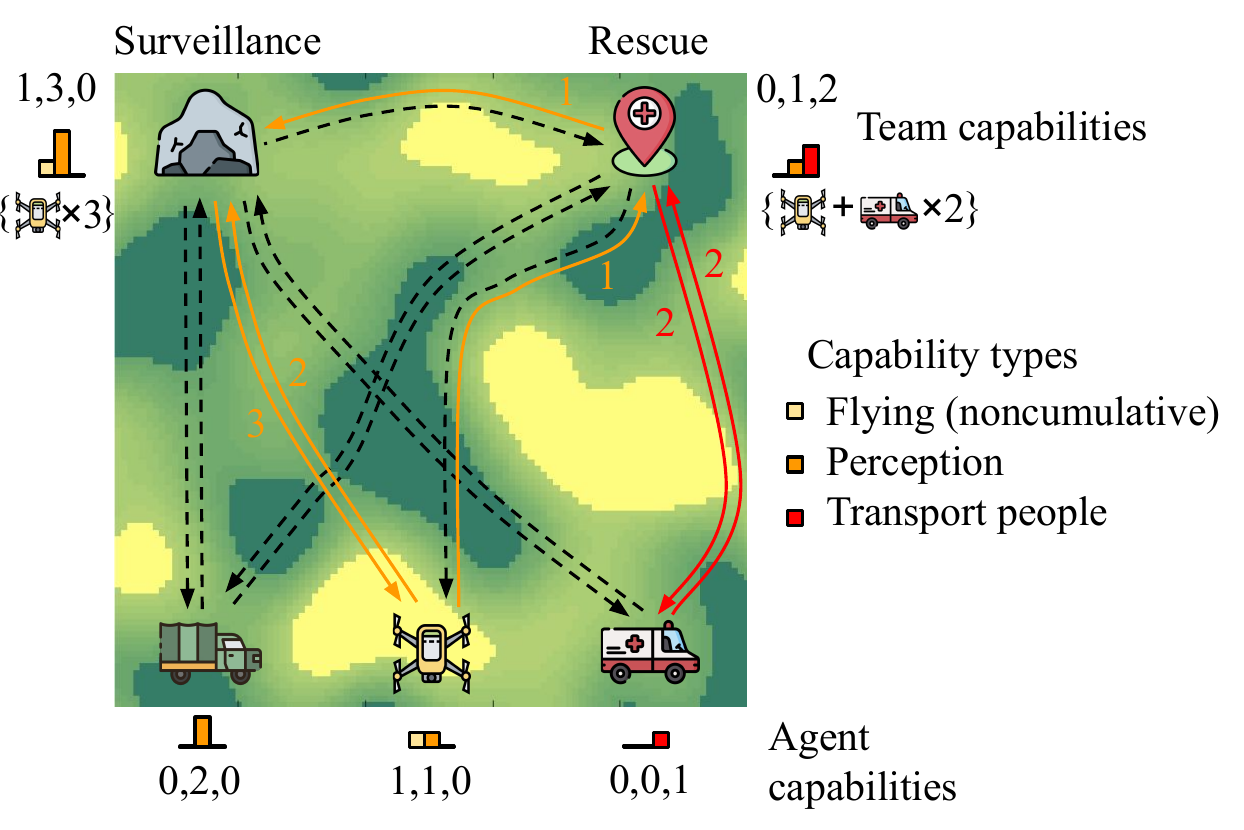}
    \caption{
    A graphical model for a problem with three agent species and two tasks. There are three types of capabilities: flying, perception, and transport people.
    The capability values are defined according to actual tasks. For instance, a quadcopter that can fly, explore an area at the rate of 1 m\(^2/\)s on average, and seat 0 people, has the expected capability values as \([1,1,0]^\transpose\).
    The rescue task requires the ability of perception and transporting people. The surveillance task requires both flying and perception capabilities.
    The capability to fly is noncumulative, which is defined in in Sec. \ref{sec:terminology} and equation \eqref{eqn:task_input}.
    The framework will consider time, energy, agent capabilities, and task constraints, and generate the schedules and routes for the teams.
    In this example, there are three agent species for selection and after the optimization, the rescue task is assigned to a team of two ambulances and a quadcopter, while the surveillance task is conducted by three quadcopters.
    }
    \label{fig:model_overview}
\end{figure}

\begin{figure*}[t]
    \centering
    \includegraphics[width=1\linewidth, trim=0 0 0 0, clip]{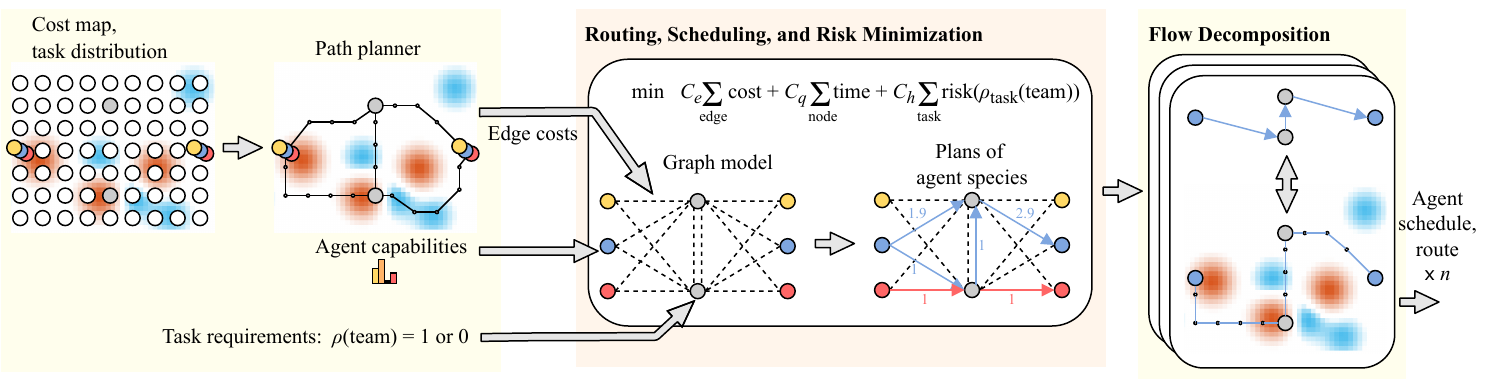}
    \caption{System architecture. Input: cost (energy) maps (an example is in Sec. \ref{sec:explore_experiment}), agent capabilities, and task requirements. Output: agent routes, schedules, and team formations. There are two major components. A routing, scheduling, and risk minimization model (Sec. \ref{sec:problem_model}) generates a network flow for each agent species. And then these networks are further decomposed into individual agent plans through a flow decomposition model (Sec. \ref{sec:flow_decompose}).}
    \label{fig:flow_chart}
\end{figure*}

The reason that a team of less powerful robots can achieve the same or more complex tasks with the above advantages is that the functional heterogeneity within the team, i.e., distinct sensor and actuator capabilities of the robots, can complement each other during a task.
In contrast to structural heterogeneity (e.g., maximum velocity or energy capacity), functional heterogeneity usually leads to fundamental differences between task capabilities (e.g., the ability to fly or transport people) \cite{sundar2017path}.
The concepts of agent and team capabilities and task requirements are described in Sec. \ref{sec:terminology}.

The fundamental problems \cite{korsah2013comprehensive} that arise when applying a functionally heterogeneous multi-agent system to a mission containing multiple complex tasks include: understanding how to decompose a task into elemental tasks (how), determining which agents should be assigned to a particular elemental task (who), and deriving a schedule that enables the heterogeneous team to successfully complete the task (when).
Therefore, considering the whole problem as task allocation; then, task allocation = task decomposition + assignment + scheduling.
The concepts of task decomposition, elemental tasks, and complex tasks are formally defined in Sec. \ref{sec:terminology}.
In the example in Fig. \ref{fig:model_overview}, the task decomposition problem considers how to decompose a task's required capabilities into agent roles such that after being assigned to the appropriate agents, the team capability surpasses the requirements. Task scheduling determines the routes such that the agents will arrive at a same task together while minimizing the energy and time usage.

A task is considered complex if there exist multiple decompositions and it is unknown which decomposition should be selected without simultaneously considering task assignment and schedule planning (i.e., the problem is coupled) \cite{korsah2013comprehensive}.

Among the few previous works that deal with functional heterogeneity in a complex task setting, most optimization models are designed for use with a specific application.
Therefore, there is a need for fundamental methods that provide a more systematic representation that can generalize within a larger scope of complex tasks.

An important aspect to consider when dealing with dynamic systems is the concept of uncertainty. Two strategies can be adopted to counter failures \cite{zhou2021multi}: 1) robust planning algorithms that consider the uncertainty pre-execution and generate a plan that is prepared to withstand the uncertain environment; 2) adaptive and reactive algorithms that recover the system from failure due to uncertainties in real-time.

In a task allocation setting, one source of the uncertainty is the inaccurate estimation of or the inherent variation within the task requirements and agent capabilities. We apply the robust planning strategy to generate more reliable plans in the pre-execution stage. A robust task allocation plan and schedule should secure task success under such uncertainty by providing redundant capabilities in the team.
Mathematically, a more robust task assignment plan means there is a higher probability that the number of capabilities in a team exceeds the number required by the task.
However, such a redundancy increases the time and energy cost, and a careful trade-off is required.
In Sec. \ref{sec:problem_model}, we will introduce how we encode this robustness in the optimization and determine a trade-off.

In this paper, we present a \textbf{Capability-based robust Task Assignment and Scheduling (\modelrisk{})} framework to optimize task decomposition, assignment, and scheduling simultaneously within a stochastic task model, which can generalize to multiple practical scenarios.
Consider a set of heterogeneous agents (can be robots/vehicles/humans) and complex tasks; the proposed framework will form agent teams, schedules, and routes to minimize energy and time costs for completing the task combined with the risk of non-completion.
We define this class of problems as the \textbf{heterogeneous teaming problem (HTP)}.
The system architecture is summarized in Fig.~\ref{fig:flow_chart} and will be discussed in the following sections.

Note that the system focuses on the optimization that distributes the capabilities to the tasks at a low cost.
The information about agent capabilities, task requirements, and spatial traveling costs are gathered and a routing and teaming plan is generated premission in a centralized planner. Individual plans are sent to the agents and do not change during the task execution.
Though the plans do not change, they are designed to withstand the uncertainty and variations in capability estimation and are less likely to fail. Future extensions will consider a recovery strategy and update the plans during the task execution. 
Communication between agents that enables information gathering and plan synchronizing is not the focus of this paper and will be considered in future work.

\subsection{Contributions}
In \cite{fu2020heterogeneous}, we dealt with a deterministic variation of such a problem, where we assumed exact information (instead of a distribution) of the agent capabilities and task requirements was known. In this work, we develop a generalizable framework for task assignment and scheduling that systematically represents heterogeneous and uncertain task requirements and agent capabilities. This paper provides the following contributions

\begin{enumerate}[label={\arabic*)}] 
  \item The development of a modeling framework that captures uncertainties within the task requirements and agent capabilities.  \label{item:first_contribution} 
  \item The derivation of a cost function that incorporates the concept of risk within the minimization.
  \item Reformulation of the heterogeneous teaming problem (HTP) to provide a more scalable algorithm that is solved by using a flow decomposition subproblem.
  \item The implementation of a capture-the-flag game simulation to compare the task assignment performance to a baseline algorithm.
\end{enumerate}

The problem size depends on the number of tasks, agent species, and agents per species. With the reformulation in this paper (compared to \cite{fu2020heterogeneous}), the routing of individual agents is decoupled and postponed to a flow decomposition subproblem which reduces the size of the optimization program for the HTP.
The HTP only considers the behavior of an entire agent species, and its size no longer depends on the number of agents per species. The complexity of the framework is still exponential, as we use mixed-integer programming to find the exact optimal solution \cite{karp1972reducibility}. But the reformulation suppresses the growth such that the framework can solve practical problems with up to 140 agents and 40 tasks.

The open-source code is available at: \\ \indent
{\small \href{https://brg.engin.umich.edu/publications/robust-task-scheduling/}{https://brg.engin.umich.edu/publications/robust-task-scheduling}}

\subsection{Outline}
The remainder of this paper is organized as follows. Sec.~\ref{sec:related_work} briefly introduces the related work of multi-agent task allocation and concludes with research gaps that this work investigates.
Sec. \ref{sec:problem_model} describes the general problem mathematically and presents a risk minimization framework.
Sec. \ref{sec:lshaped_algorithm} introduces an algorithm to solve the risk minimization problem and output a general routing plan described as network flows of agent movements.
Sec. \ref{sec:flow_decompose} proposes a network flow decomposition problem to generate routing plans and schedules for individual agents.
The two-step process in Sections \ref{sec:lshaped_algorithm} and \ref{sec:flow_decompose} improves the scalability of the framework compared to a solution method that outputs individual plans directly within one optimization.
Sec.~\ref{sec:experiments} discusses the experiments and results.
Finally, Sec.~\ref{sec:conclusion} concludes the paper and provides ideas for future work based on current limitations.

\subsection{Concepts and Terminology}\label{sec:terminology}

In this section, we provide definitions of multiple concepts that will be used throughout the paper. For some of the terminology, we refer to \cite{korsah2013comprehensive}, but simplify the description to provide only the core meaning.

\textbf{Agent capabilities.} A numeric vector that describes the agent's fitness to conduct a task.

\textbf{Team capabilities.} A set of fitness values aggregated from agent capabilities in the team.

\textbf{Cumulative capabilities.} A capability that sums the agent capabilities across an entire team to get the team capability.

\textbf{Noncumulative capabilities.} A capability that does not sum.
In this paper, noncumulative team capability is the minimum of the agent capabilities in the team.\footnote{In Fig. \ref{fig:model_overview}, the capability to fly is noncumulative: the team can fly only when all agents in the team can fly.}

\textbf{Task requirement.} A set of constraints that relate the task and agents. Usually, it is specified as a number of required agents, agent actions, or agent capabilities that have to be satisfied. In this paper, it is defined as a set of required team capabilities and will be described in Sec. \ref{sec:problem_model}.

\textbf{Task decomposition.} Representing a task as a set of elemental tasks (there can be constraints between these subtasks), such that completing the subtasks completes the original task.

\textbf{Task assignment.} Identifying which specific agent should handle an elemental task.

\textbf{Elemental task}: A task that is not decomposable and should be assigned to only one agent.

\textbf{Compound task.} A task that can be decomposed into a set of elemental tasks in only one fixed way.

\textbf{Complex task.} A task that can be decomposed into elemental tasks in more than one way, i.e., following different decompositions, the resulting sets will contain different numbers or types of elemental tasks.

%% file: section/related_work.tex
\section{Related Work}\label{sec:related_work}

According to the taxonomy in \cite{korsah2013comprehensive, gerkey2004formal, khamis2015multi, nunes2017taxonomy, rizk2019cooperative, prorok2021beyond}, our framework for the HTP deals with a task allocation problem in the category CD-[ST-MR-TA]: complex task dependencies (CD), single-task robot (ST), multi-robot tasks (MR), and time-extended assignment (TA). It is one of the most challenging task assignment categories and has been considered in few previous works in literature. The time-extended assignment considers scheduling, in contrast to instantaneous assignments (IA), which only make matches between tasks and robots.

While many previous works about multi-robot tasks are in the category [ST-MR-TA], most of their papers do not deal with complex tasks.
Their papers decompose a multi-robot task into elemental tasks that can be assigned to a subset of single agents.
If such a decomposition exists, the task decomposition problem is presolved and decoupled from the assignment and scheduling problem. Many previous works in vehicle routing \cite{sundar2017path, korsah2012xbots}, job shop scheduling \cite{ozguven2012mixed, ku2016mixed, moser2020flexible}, and robotic soccer games \cite{liemhetcharat2011modeling, liemhetcharat2012weighted} deal with such non-complex (or compound) tasks.
We refer interested readers to the survey paper \cite{korsah2013comprehensive} for more work with compound tasks.

There are systems dealing with complex tasks, but their models do not generalize and are applied to one specific problem. Examples include fire extinguishing and debris removal \cite{jones2011time, pujol2015efficient}, and a coverage problem of unmanned aerial/ground vehicles with recharging behaviors \cite{yu2019coverage}, where tasks are considered complex because multiple decomposition combinations are possible in the problem space.

Some frameworks for CD-[ST-MR-TA] created in previous work generalize to a type of problem, but most of them have scalability issues and are only applicable to a small group of robots.
For instance, Bayesian network representations consisting of tasks, observations, constraints, and action nodes can be used to represent a task with interdependent elemental subtasks \cite{song2010learning, ek2010task}.
Hierarchical tree networks (HTN) are used in \cite{zlot2003multirobot, zlot2003market, zlot2005complex, hayes2016autonomously} to represent tasks where leaf nodes are roles for assignment.
Planning domain definition language (PDDL) can be used to represent a task as a graph of states (nodes) and actions (edges) \cite{klee2015graph, nicolescu2003natural, ekvall2008robot, niekum2012learning, grollman2010incremental, hayes2015effective, galindo2008robot, aeronautiques1998pddl, torreno2017cooperative}.
Temporal planning can be considered an extension to the basic PDDL \cite{benton2012temporal, sapena2016parallel, carreno2020decentralised, crosby2014temporal, cashmore2018temporal}. In addition to a graph of predefined actions and state-space representation, temporal planning allows continuous temporal constraints and interdependence between subtasks to be encoded as predefined logical formulas.
However, these representations based on actions and states are usually applied to systems with less than five agents. The number of states in their trees/graphs could explode rapidly with respect to the number of agents and available actions. The underlying solver (e.g., graph search algorithms) would take a long time to explore a subset of the high-dimensional space. Such a scalability issue limits its application to larger multi-agent systems.

Uncertainty exists widely in the estimation of task requirements and agent capabilities. Some recent works capture such uncertainty explicitly in their models. However, the metrics they optimize do not necessarily result in a high-probability task requirement satisfaction (which is considered robust).
For example, the work of \cite{faruq2018simultaneous} models the agent action uncertainty in Markov decision processes (MDP) and optimizes the expected objective.
In \cite{liemhetcharat2011modeling, liemhetcharat2012weighted, ravichandar2020strata}, an agent's capabilities are represented as a Gaussian random vector where the uncertainty is captured in the distribution. The work of \cite{ravichandar2020strata} then penalizes the variance of the assigned agent capabilities to limit the uncertainty.
However, expectation or variance are not well-justified quantitative metrics in such an optimization problem
and can result in problematic solutions. For instance, if the required capability of a task is matched precisely, it is reasonable to limit the variances within a small threshold. However, if the team's capability surpasses the task requirement by a lot, larger variances are acceptable. As another example, optimizing an expected cost (average performance) can be problematic for safety-critical applications.
In \cite{rudolph2021desperate}, the authors directly maximize the probability that enough capabilities are assigned. However, probability is non-convex in the decision variables, and global optima are hard to obtain for large problem cases.

\begin{figure}[t]
    \centering
    \includegraphics[width=0.7\linewidth]{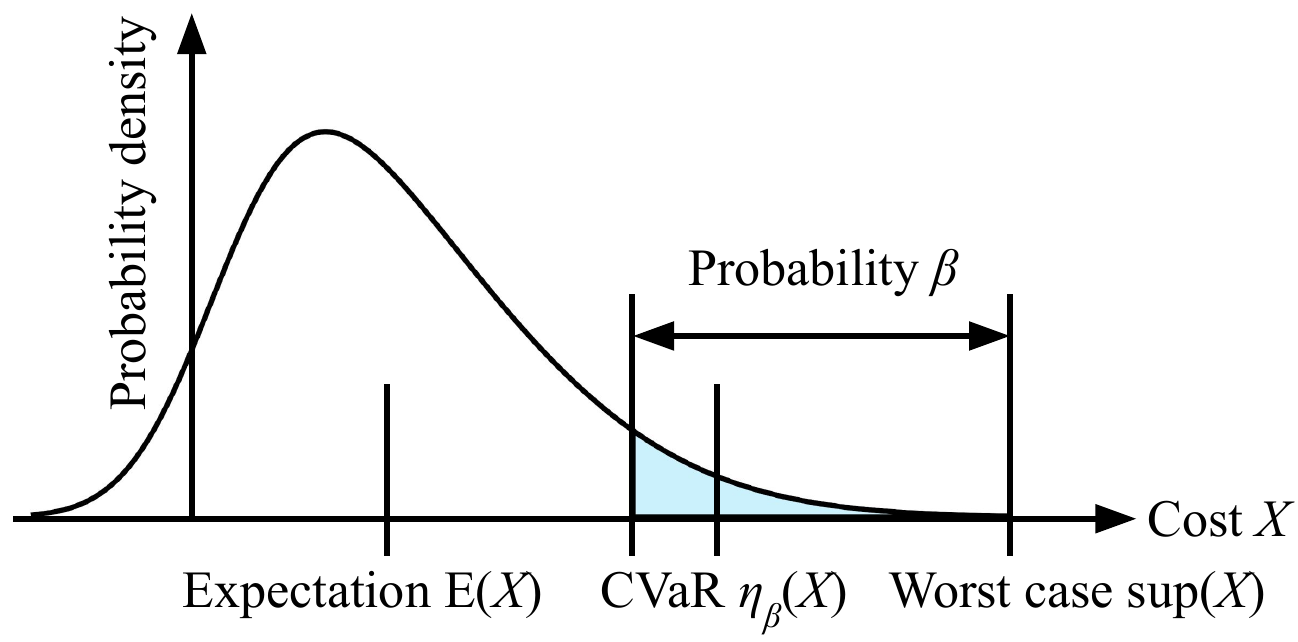}
    \caption{Graphical illustration of CVaR, defined as expected cost of the worst \(\beta\)-proportion of the cases. Denoted as \(\eta_\beta(\cdot)\), it is a function of the random distribution with \(\beta\) as a hyper-parameter.}
    \label{fig:cvar_definition}
\end{figure}

Our proposed model considers a generalizable framework for allocating agent capabilities for complex tasks under capability and requirements uncertainty, and it can be applied to systems consisting of hundreds of agents.
We represent the agent capabilities as a vector of random distributions (not necessarily Gaussian) and the task requirements within a function of the team capabilities (defined in Sec. \ref{sec:problem_model}).
The task requirement is verified once the aggregated capabilities of the team drive the binary function to one.
We then solve the task decomposition, assignment, and scheduling problem simultaneously, where we optimize the time, energy, and robustness of task allocation.
As mentioned in the introduction, a robust allocation should ensure the capability of the team exceeds the task requirement with a high probability.
We choose to minimize the conditional value at risk (CVaR) of (requirement \(-\) capability), which is consistent\footnote{According to \cite{sarykalin2008value},  maximizing the probability is equivalent to minimizing the Value at Risk, and the Conditional Value at Risk is a risk metric with additional math properties that facilitate optimization problems.}
with maximizing the probability \(P(\textnormal{capability} \geq \textnormal{requirement})\) \cite{sarykalin2008value}.
CVaR is a provably sensible measurement of the uncertainties in practical applications \cite{majumdar2020should}. It is widely accepted by the finance community and appears with a growing frequency in recent robotic applications for risk-aware single-robot control \cite{balasubramanian2020risk, bernhard2019addressing, lindemann2020control, hakobyan2019risk} and multi-robot coordination \cite{zhou2018approximation, zhou2021multi}. The definition of CVaR is illustrated in Fig. \ref{fig:cvar_definition}.

\begin{table}[t]
  \centering
  \caption{A comparison between STRATA and \modelrisk{}.}
  \label{tab:strata_vs_ours}
    \begin{tabular}{l|cc}
    \toprule
          & STRATA\cite{ravichandar2020strata} & \modelrisk{} (Ours) \\
    \midrule
    Agent capabilities & continuous/Gaussian & continuous/stochastic \\
    Capability types & cumulative & cumulative/noncumulative \\
    Task requirements & and   & and/or \\
    Scheduling & no    & yes \\
    Uncertainty control & limit variance & minimize CVaR \\
    Optimization type & continuous nonlinear & mixed-integer nonlinear \\
    \bottomrule
    \end{tabular}%
\end{table}%

Our problem can be considered a coalition formation with spatial and temporal constraints. Multiple models have been proposed in previous work \cite{koes2005heterogeneous, ramchurn2010coalition, vig2006multi}. The work of \cite{ramchurn2010coalition} has a similar application to ours, but it assumes that there exists a predefined utility function that outputs a value for a team configuration. While an abstract function generalizes the problem space, explicitly defining the function values for all team configurations could be hard in practice. We provide our task requirement and agent capability model to avoid such an explicit definition.
The work of \cite{neville2021interleaved} and its extension \cite{messing2022grstaps} use a similar requirement and capability model. However, they do not consider the uncertainty in their capability model. A graph search-based algorithm is applied in their task allocation process, limiting the scalability of the number of agents (the size of the graph).

STRATA \cite{ravichandar2020strata, prorok2017impact} shares many similarities with our proposed approach, \modelrisk, including stochastic agent capability vectors and task requirements on the team's capabilities. Therefore, we choose STRATA as our baseline algorithm in one of the experiments in Sec. \ref{sec:experiments}.
However, STRATA falls in the area of CD-[ST-MR-IA], where it assumes the tasks happen at the same time, which simplifies the scheduling problem.
Meanwhile, our task model, represented using requirement functions, is more expressive. TABLE \ref{tab:strata_vs_ours} provides a more detailed comparison between the two models.
The meaning of `and/or' in the table is defined in Sec. \ref{sec:problem_model}. Note that the definition of `noncumulative' in STRATA differs from ours. STRATA claims that it can deal with noncumulative capability types (such as the speed of an agent). However, it thresholds on a value and then treats the binary value after thresholding as a cumulative capability. For noncumulative types, we define the team's capability to be the minimum of all agents and require instead that all agents in the team meet the minimum task requirement. STRATA is not able to enforce such requirements for their noncumulative capabilities.

%% file: section/model.tex
\section{Routing, Scheduling, and Risk Minimization}\label{sec:problem_model}

In this section, we formally define the heterogeneous teaming problem with uncertain agent capabilities and task requirements. We describe the task requirement functions and agent capability vectors used to represent the task structure, organize them in a graphical model, and then encode the task allocation problem as a stochastic mixed-integer program whose objective jointly consists of time, energy, and risk cost.

The modeling assumptions are listed as follows:
\begin{itemize}
    \item Tasks are distributed spatially and can be completed after a known time once a team with the required capabilities arrives.
    \item The agent capabilities can be modeled as a vector of known continuous random distributions.
    \item The task requirement can be verified by a binary function of the agent team's capabilities. The parameters in the function are task-required capabilities and are known continuous random distributions.
    \item Different types of capabilities are independent and do not affect each other.
    \item The completion of one task does not depend on other tasks. But the scheduling dependency between tasks is considered, i.e., agents should arrive at the same time for a task, and the delay of one task could delay the subsequent tasks. The model can be extended with other time constraints, e.g., precedence constraints.
\end{itemize}

\subsection{Heterogeneous Teaming Problem Description}\label{sec:problem_description}
We simplify the graphical model in \cite{fu2020heterogeneous} for this work.
Consider a set of agent species \(V = \{1,\cdots,{n_v}\}\), capability types \(A = \{1,\cdots,{n_a}\}\), and tasks \(M = \{1, \cdots, {n_m}\}\). 
Note that \(V, A, M \subset \mathbb{N}\) are three integer sets.
Each agent species \(k \in V\) is associated with a non-negative capability vector \(\mathbf{c}_k = [c_{k 1}, \cdots, c_{k {n_a}}]^\transpose\), where \(c_{k a}\) is a random variable with a known distribution, representing the uncertain task capability of type \(a \in A\) for agent \(k \in V\). Each task \(i \in M\) requires an agent team with appropriate capabilities that drives a task requirement function \(\rho_i(\cdot)\) to 1. 

A task requirement function is a binary function of a similar structure as \eqref{eqn:task_requirement_function}. The logical operators \(\geq\), \(\land\), and \(\lor\) are `greater than or equal to', `and', and `or' that return 1 if their conditions are satisfied, and 0 otherwise. \(\gamma_{1 a}\) describes the task \(1\)'s requirement on capability \(a\), and is modeled as a random variable with a known distribution. We define \(\gamma_{1 a} = 0\) if it does not appear in \eqref{eqn:task_requirement_function}. Note that \eqref{eqn:task_requirement_function} is an example of a task that requires four types of capabilities. In practice, there can be an arbitrary number of \(\land\) and \(\lor\), theoretically. 
\begin{align}
    & \rho_{1}(\alpha_{1}, \alpha_{2}, \cdots) = [(\alpha_{1} \geq \gamma_{1 1}) \lor (\alpha_{2} \geq \gamma_{1 2})] \nonumber \\
    & \quad\quad\quad\quad\quad\quad\quad \land [\alpha_{3} \geq \gamma_{1 3}] \land[\alpha_{4} \geq \gamma_{1 4}]. \label{eqn:task_requirement_function} \\
    & \alpha_a =
    \begin{cases} {\sum}_{k \in V} c_{k a} \cdot y_{k 1}, \ a \text{ is cumulative} \\
    {\min}_{k \in V} c_{k a} \cdot r_{k 1}, a \text{ is noncumulative}
    \end{cases}\hspace{-0.1in}, 
    \forall a \in A.
     \label{eqn:task_input}
\end{align}

\(\alpha_a\) is the capability \(a\) of the agent team.
\(y_{k i}\) is the number of agents at task \(i \in M\) with species \(k \in V\). \(r_{k i}\) is binary and indicates whether there is an agent with species \(k \in V\) at task \(i \in M\). Depending on whether the capability is cumulative, we can compute \(\alpha_a\) according to \eqref{eqn:task_input}. An example of noncumulative capabilities is the speed limit of a team. It equals the speed of the slowest moving agent in the team.

The requirement to drive these functions to 1 could be satisfied with appropriate team formation planning. This requirement constraint can be encoded as linear constraints according to \cite{fu2020heterogeneous}. Note that the only part that could introduce non-convexity is the logic \(\lor\), which takes the union of two feasible regions. Reducing the number of \(\lor\) operators is desired as it facilitates the optimization.

With stochasticity in task requirements and agent capabilities, the goal is to determine the optimal task schedule for a selected set of agents, such that the energy, time, and path constraints are satisfied, and the energy cost and risk of the task's non-completion are jointly minimized.

The task is considered complex for two reasons. First, the logic \(\lor\) operator results in multiple decompositions. Secondly, there is no fixed way to decompose the requirements in \eqref{eqn:task_requirement_function} into single-agent elemental tasks; instead, they are optimized together with the assignment and scheduling problem.

There are inter-schedule dependencies between tasks since the overall time cost for one agent to achieve one task depends on the schedule of other agents. For example, if one agent delays its work at another task and arrives late at the current task, the time cost increases for all agents at the current task.

\subsection{Routing Model}
\begin{figure}[hbt!]
	\centering
	\includegraphics[width=0.9\linewidth]{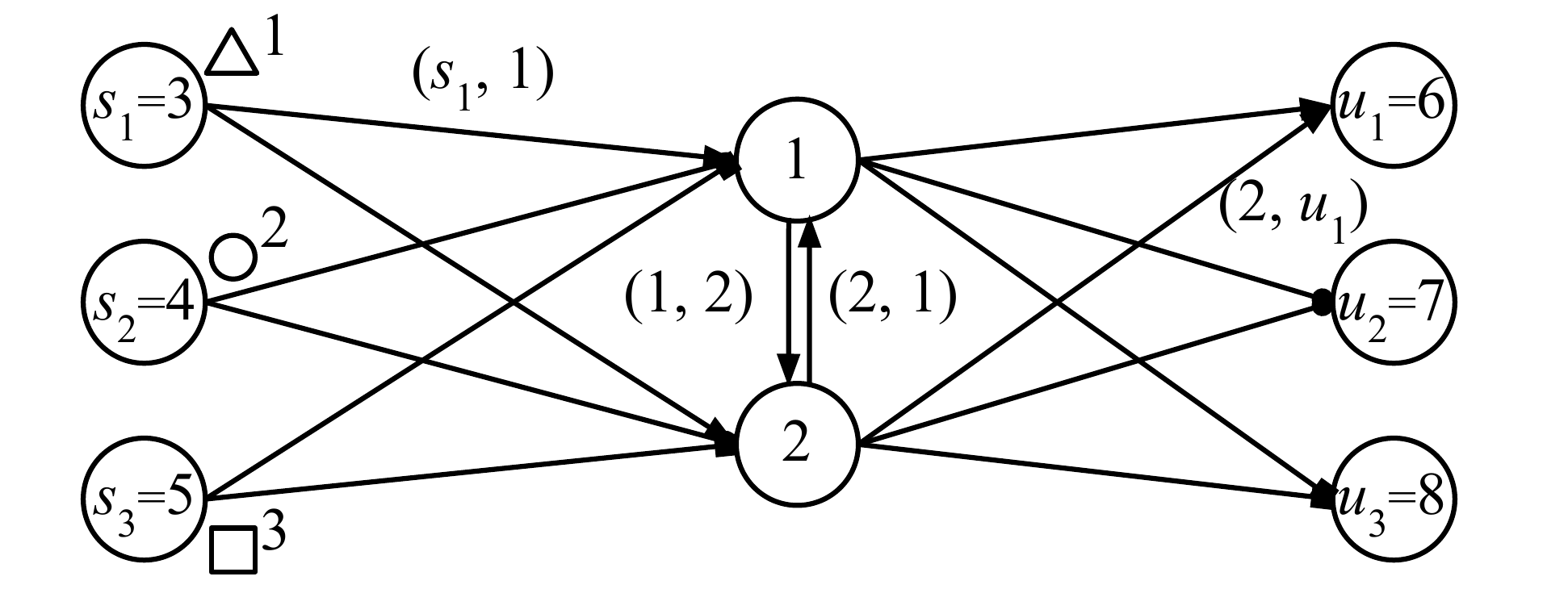}
	\caption{Graphical model with \(n_v = 3\) agent species and \(n_m = 2\) tasks.
	}
	\label{fig:graphical_model}
\end{figure}

As shown in Fig. \ref{fig:graphical_model}, we first define a directed graph \mbox{\(G = (N,E)\)}, with a set of vertices \(N = S \cup U \cup M\) and edges \(E\). \(M = \{1, \cdots, {n_m}\}\) is the set of task nodes. \(S = \{n_m + 1, \cdots, n_m + n_v\}\) and \(U = \{n_m + n_v + 1, \cdots, n_m + 2 n_v\}\) are the sets of start and terminal nodes for each agent species \(k \in V\).
The prefix \(n_m\) and \(n_m + n_v\) are added such that the vertex indices in \(M\), \(S\), and \(U\) are unique integers.
The size of the vertices set \(M\), \(S\), and \(U\) are \(n_m\), \(n_v\), and \(n_v\), respectively. Note that these nodes can represent the same or different physical locations in the real world.
A practical example of this graphical model is in Fig. \ref{fig:model_overview}.

The edge set \(E = \{(i,j) | \ \forall i \in S \cup M, \ j \in M \cup U\}\): there are edges from the starts to all tasks, from all tasks to the terminals, and between all the tasks.
The edges are associated with their time and energy costs and the pre-computed real-world paths between two node locations. There is an edge between every task node pair in Fig. \ref{fig:graphical_model}, but in practice, we only add an edge if there is a feasible path between two nodes.

Under this setting, the agents of species \(k \in V\) should start at \(s_k\), follow the edges to visit a subset of task nodes (often together with other agents), and terminate at \(u_k\). As a result, the agent numbers on the edges form a network flow from the start to the terminal nodes in the given graph \(G\).

\subsection{Risk Minimization Model - A Mixed-Integer Program}

We discuss an MIP model that generates a schedule and agent team for each task and the corresponding agent flow by minimizing the energy cost and the risk of the task's non-completion.
Here we provide common notations that will be used in TABLE \ref{tab:variable_definition}. Note that the decision variables are \(x_{kij}\), \(y_{ki}\), \(r_{kij}\), \(r_{ki}\), \(q_i\), and \(g_{k i}\). 

\begin{table}[t]
  \caption{Definition of the notation.}
  \label{tab:variable_definition}%
    \begin{tabular}{p{0.06\linewidth}|p{0.8\linewidth}} 
    \toprule
     & Meaning
    \\
    \midrule
    \(x_{k i j}\)& The number of agents on edge \((i, j)\) with species \(k \in V\), where node \(i, j \in N\).
    For simplicity, \(x_{k s_k i}\) and \(x_{k i u_k}\) \(\forall k \in V, i \in N\) are abbreviated to  \(x_{k s i}\) and \(x_{k i u}\) since there is no ambiguity.
    The colored numbers in Fig. \ref{fig:model_overview} are solutions to \(x_{k i j}\) for that example problem.
    \\
    \(y_{k i}\)& The number of agents at task \(i \in M\) with species \(k \in V\).
    \\
    \(r_{k i j}\)& = 1, if \(x_{k i j} \geq 1\), otherwise 0. (A helper variable indicating whether there are agents with species \(k\) on that edge.)
    \\
    \(r_{k i}\)& = 1, if \(y_{k i} \geq 1\), otherwise 0. (A helper variable indicating whether there are agents with species \(k\) at that task.)
    \\
    \(q_i\)& The time that task \(i \in M\) begins.
    \\
    \(g_{ki}\)& The maximum cumulative energy that an agent of species \(k \in V\) has spent at node \(i\). (A helper variable.)
    \\
    \(b_{k i j}\)& The deterministic energy cost for agent \(k \in V\) to travel edge \((i, j) \in E\).
    \\
    \(t_{k i j}\)& The deterministic time for \(k \in V\) to travel edge \((i,j) \in E\).
    \\
    \(t_{k i}\)& The deterministic time for agent species \(k \in V\) to complete its part for task \(i \in M\).
    \\
    \(C_{q}\)& Time penalty coefficient.
    \\
    \(C_{\text{large}}\)& A large constant number for the MIP.
    \\
    \(B_k\)& The energy capacity of agent \(k \in V\).
    \\
    \(B_{\text{large}}\) & A large constant energy for the MIP.
    \\
     \(h_i\)& The conditional value at risk from task \(i \in M\).
    \\
    \(n_k\) & The number of agents with species \(k \in V\).
    \\
    \bottomrule
    \end{tabular}
\end{table}

Compared to \cite{fu2020heterogeneous}, the \(x_{kij}\) and \(y_{ki}\) variables in this work are defined for agent species instead of individual agents, and their values are now real instead of binary. This change reduces the number of variables and, therefore, decreases the computational cost of the planner. However, as the model does not specify variables for individual agents, the mixed-integer program (MIP) outputs an agent flow of each species instead of path plans for individual agents. We will discuss algorithms for post-processing the agent flows to get routes for individual agents in Sec. \ref{sec:flow_decompose}.

\subsubsection{Variable Bounds}
\begin{align}
    \begin{split}
    &x_{kij} \geq 0, \ y_{ki} \geq 0, \quad \forall i,j \in \node, \forall k \in V. \\
    &r_{kij}, \ r_{ki}\in \{0,1\}, \quad \forall i,j \in \node, \forall k \in V. \\
    &q_i \geq 0, \quad \forall i \in M \cup U. \quad q_i = 0, \quad \forall i \in S. \\
    &g_{k i} \geq 0, \quad \forall i \in M \cup U. \quad g_{k i} = 0, \quad \forall i \in S, \forall k \in V.
    \end{split}\label{eqn:bound_constraint}
\end{align}

The numbers of agents from species \(k \in V\) on an edge and at a task node are positive and real. Their helper variables are binary. The time and cumulative energy are nonnegative at all nodes and zero at the start nodes. \\

\subsubsection{Helper-variable Constraints}
\begin{align}
    \begin{split}
    &x_{kij} \geq r_{kij}, \quad x_{kij} \leq n_k \cdot r_{kij}, \quad \forall i,j \in \node, \forall k \in V. \\
    &y_{ki} \geq r_{ki}, \quad y_{ki} \leq n_k \cdot r_{ki}, \quad \forall i \in \node, \forall k \in V.
    \end{split}\label{eqn:x_r_constraint}
\end{align}    

Equation \eqref{eqn:x_r_constraint} encodes the relationship between \{\(x_{kij}\), \(y_{ki}\)\} and their helper variables \{\(r_{kij}\), \(r_{ki}\)\}. \\

\subsubsection{Network Flow Constraints}
\begin{align}
    \underset{i \in S \cup M} {\sum} x_{k i m} &= \underset{j \in U \cup M} {\sum} x_{k m j}, \quad \forall m \in M, \ \forall k \in V. \label{eqn:flow_constraint1} \\
    \underset{i \in M} {\sum} x_{k s i} &\leq n_k, \quad \quad \quad \forall k \in V. \label{eqn:flow_constraint2} \\
    y_{k j} &\leq \underset{i \in M} {\sum} x_{k s i}, \quad \forall j \in M, \ \forall k \in V. \label{eqn:flow_constraint3} \\
    y_{k j} &= \underset{i \in S \cup M} {\sum} x_{k i j}, \quad \forall j \in M, \ \forall k \in V. \label{eqn:node_vehicle_constraint}
\end{align}

Equation \eqref{eqn:flow_constraint1}-\eqref{eqn:flow_constraint3} are flow constraints that ensure the agent numbers are smaller than the upper bound, and that the incoming agent number at a node equals the outgoing number. Constraint \eqref{eqn:node_vehicle_constraint} reflects the relationship that the number of agents at a node equals the sum of the agent flows from all incoming edges. \\

\subsubsection{Energy Constraints}

\begin{align}
    g_{k i} - g_{k j} + b_{k i j} &\leq B_{\text{large}} (1 - r_{k i j}),
    \forall i , j \in \node, \forall k \in V. \label{eqn:energy_constraint1} \\
    g_{k i} &\leq B_{k}, \quad \forall i \in \node, \ \forall k \in V. \label{eqn:energy_constraint}
\end{align}

Equation \eqref{eqn:energy_constraint1} ensures that \(g_{k i}\) is the cumulative energy of the agent species - if an edge is traveled, the energy \(b_{k i j}\) is added.
And because \(g_{k i}\) is the maximum cumulative energy of agent species \(k\) at node \(i\), equation \eqref{eqn:energy_constraint} ensures the energy cost for an agent of species \(k\) does not exceed its energy capacity \(B_k\).

These energy constraints form sufficient conditions as they assure the most costly path in the flow network of species \(k\) is within the capacity limit \(B_k\). However, it is possible that the most costly path is not picked during the flow decomposition procedure (Sec. \ref{sec:flow_decompose}). This is a compromise made by us along the process of improving the scalability of the model: we replace the variables for individual agents with variables for an entire agent species. With the original variables for individual agents, necessary and sufficient energy constraints were easily imposed. \\

\subsubsection{Time Constraints}
\begin{align}
    q_{i} - q_{j} + t_{k i j} + t_{k i} &\leq C_{\text{large}} (1 - r_{k i j}),
    \forall i , j \in \node, \forall k \in V. \label{eqn:time_constraint1}
\end{align}

Equation \eqref{eqn:time_constraint1} is a scheduling constraint: for an agent, the time duration between two consecutive tasks should be larger than the service time at the previous task plus the travel time.
This constraint gives all the agents in the team enough time to arrive before the current task starts.

With the task time variables \(q_{i}\), the optimization problem can be extended with other user-defined time constraints. For instance, precedence constraints between tasks or time window constraints within which a task should complete \cite{fu2021simultaneous}.
\\

\subsubsection{Task Requirement Constraints}
\begin{align}
    & 1 = \rho_i(\underset{k \in V}{\sum} E(c_{k 1}) y_{k i}, \underset{k \in V}{\sum} E(c_{k 2}) y_{k i}, \ \cdots), \ \forall i \in M.  \label{eqn:task_complete_constraint1} \\
    & 1 = \rho_i(\underset{k \in V}{\min} E(c_{k 1}) r_{k i}, \underset{k \in V}{\min} E(c_{k 2}) r_{k i}, \ \cdots), \ \forall i \in M.  \label{eqn:task_complete_constraint2}
\end{align}

For cumulative capabilities we add constraint \eqref{eqn:task_complete_constraint1}, and for noncumulative capabilities we add constraint \eqref{eqn:task_complete_constraint2}. \(E(\cdot)\) is the expectation operator. 
\(c_{ki}\) is the capability value defined in Sec. \ref{sec:problem_description}.
Note that an example of \(\rho_i(\cdot)\) is in \eqref{eqn:task_requirement_function}, and \(\gamma_{i a}\) \((\forall i \in M, a \in A)\) should be replaced with \(E(\gamma_{i a})\) here.
Though the task requirements and the agents' task capabilities are stochastic, we can add this deterministic constraint that requires the teams' expected capabilities to satisfy the task requirement by driving the requirement function to 1. \\

\subsubsection{Objective Function}
\begin{align}
    \min &\ C_e \underset{k \in V}{\sum} \ \underset{i,j \in \node}{\sum} b_{k i j} \cdot x_{k i j} + C_{q} \underset{i \in U}{\sum} q_i 
    + C_h \underset{i \in M}{\sum} h_i
    \label{eqn:nonlin_penalty_objective} \\
    h_{i} &= \sum_{a \in A \text{ s.t. } \gamma_{i a} \neq 0 } h_{i a}, \quad \quad \quad \quad \quad \quad \quad \quad \ \ \ \forall i \in M, \label{eqn:nonlin_cvar_objective_h} \\
    h_{i a} &= \underset{k \in V \text{ s.t. }r_{k i} = 1}{\max} \eta_\beta \left({-c_{k a} + \gamma_{i a}} \right), \ \ a \in A, \forall i \in M,  \label{eqn:nonlin_cvar_objective_h_min} \\
    h_{i a} &= \eta_\beta ( {-\underset{k \in V}{\sum} c_{k a} \cdot y_{k i} + \gamma_{i a}} ), \quad \quad \   a \in A, \forall i \in M. \label{eqn:nonlin_cvar_objective_h_sum}
\end{align}

In the objective function \eqref{eqn:nonlin_penalty_objective}, we want to minimize a weighted combination of the energy cost, task time, and the conditional value at risk of the task's non-completion, where \(C_e\), \(C_q\), and \(C_h\) are user-defined weights.
For the risk of a task, \(h_i\) in \eqref{eqn:nonlin_cvar_objective_h}, take task 1 and its example requirement function in \eqref{eqn:task_requirement_function} as an instance. According to \eqref{eqn:task_requirement_function} and \eqref{eqn:nonlin_cvar_objective_h}, \(h_{1} = h_{1 1} + h_{1 2} + h_{1 3} + h_{1 4}\).
To explain this, task \(1\) requires capabilities \(1\)-\(4\). Assuming the requirements on these capabilities are independently placed, the total risk is the sum of the risks from each requirement.

Consider \(h_{i a}\), the risk on a single capability \(a \in A\).
According to the example requirement function \eqref{eqn:task_requirement_function}, a task requests \(\alpha_{a} \geq \gamma_{i a}\). Both \(\alpha_{a}\) and \(\gamma_{i a}\) are stochastic. To maximize the probability \(P(\alpha_{a} \geq \gamma_{i a})\), we instead minimize the CVaR of \(\gamma_{i a} - \alpha_{a}\). Let the function \(\eta_\beta(\cdot)\) be the CVaR of a random variable with probability level \(\beta\). Then, \(h_{i a}\) can be computed according to \eqref{eqn:nonlin_cvar_objective_h_sum} or \eqref{eqn:nonlin_cvar_objective_h_min} depending on whether capability \(a \in A\) is cumulative or not.

Given the uncertainty in the capabilities, the objective function \eqref{eqn:nonlin_penalty_objective}, together with the deterministic task requirement constraint \eqref{eqn:task_complete_constraint1}, tries to maximize the probability of task success at a low energy and time cost.

\subsection{Sample Average Approximation (SAA)}
Solving the optimization specified by \eqref{eqn:bound_constraint}-\eqref{eqn:nonlin_cvar_objective_h_sum} requires dealing with stochasticity and nonlinearity. In this section, we show how to convert this stochastic mixed-integer nonlinear program (MINLP) to a deterministic mixed-integer linear program (MILP).

Notice that \(E(c_{k a})\) in \eqref{eqn:task_complete_constraint1} and \(\eta_\beta(-c_{k a} + \gamma_{i a})\) in \eqref{eqn:nonlin_cvar_objective_h_min} do not involve decision variables, these can be computed prior to the optimization, given the distribution of \(c_{k a}\) and \(\gamma_{i a}\) (\(k \in V, a \in A\)). The stochasticity is eliminated by the expectation and risk function. The deterministic task requirements in \eqref{eqn:task_complete_constraint1} can then be represented with a set of linear constraints according to \cite{fu2020heterogeneous}. Constraint \eqref{eqn:nonlin_cvar_objective_h_min} can also be represented as a linear constraint
\begin{align}
    h_{i a} \geq \eta_\beta \left({-c_{k a} + \gamma_{i a}} \right) \cdot r_{k i}, \quad \forall k \in V, a \in A, \forall i \in M. \label{eqn:lin_cvar_objective_h_min}
\end{align}

The only non-linearity is in \eqref{eqn:nonlin_cvar_objective_h_sum} due to the function \(\eta_\beta(\cdot)\) and the decision variable \(y_{k i}\). However, we can linearize \eqref{eqn:nonlin_cvar_objective_h_sum} using the sample average approximation algorithm. Suppose we can represent the random distribution \(c_{k a}\) and \(\gamma_{i a}\) by samples \(c_{k a}^{(\xi)}\) and \(\gamma_{i a}^{(\xi)}\) (\(\xi = 1, \cdots, n_\xi\)), respectively. Then we can approximate \eqref{eqn:nonlin_cvar_objective_h_sum} with a linear equation \eqref{eqn:cvar_objective_h_sum_linear} according to \cite{rockafellar2000optimization}.
The approximation converges at the rate of \(\bigO{n_\xi^{-1/2}}\) \cite{asmussen2007stochastic}.
In the following equation, \(\lambda_{i a}\) is a continuous helper variable for calculating \(h_{i a}\) 
(task \(i \in M\) and capability type \(a \in A\)).
\begin{align}
    h_{i a} = & \lambda_{i a} + {1} / {n_\xi (1 - \beta)} \cdot \nonumber \\
    & \stackrel{n_\xi}{\underset{\xi = 1}{\sum}}
    \left[
    -\underset{k \in V}{\sum} c_{k a}^{(\xi)} \cdot y_{k i} + \gamma_{i a}^{(\xi)} - \lambda_{i a}
    \right]^{+}. \label{eqn:cvar_objective_h_sum_linear} \\
    [x]^+ =& \begin{cases} x, \quad x > 0 \\
    0, \quad x \leq 0
    \end{cases}. \label{eqn:relu_function}
\end{align}

Finally, the piece-wise linear function in \eqref{eqn:cvar_objective_h_sum_linear} can be represented as linear constraints in \eqref{eqn:cvar_objective_h_sum_linear2}. Note that we add a series of continuous helper variables \(w_{i a}^{(\xi)}\) to represent the \([\cdot]^+\) function in \eqref{eqn:relu_function}.
\begin{align}
    h_{i a} &= \lambda_{i a} + \frac{1}{n_\xi (1 - \beta)} \stackrel{n_\xi}{\underset{\xi = 1}{\sum}} w_{i a}^{(\xi)}, \nonumber \\
    w_{i a}^{(\xi)} &\geq -\underset{k \in V}{\sum} c_{k a}^{(\xi)} \cdot y_{k i} + \gamma_{i a}^{(\xi)} - \lambda_{i a}, \label{eqn:cvar_objective_h_sum_linear2} \\
    w_{i a}^{(\xi)} &\geq 0. \nonumber
\end{align}

%% file: section/optimization_algorithm.tex
\subsection{The L-shaped Algorithm}\label{sec:lshaped_algorithm}

In the previous section, we formulate the stochastic heterogeneous teaming problem as an MINLP optimization and approximate it as an MILP using sample average approximation. However, the number of variables and constraints in the linear program is a function of the sample number \(n_\xi\). When the sample number dominates, the size of the linear program is roughly \(\bigO{n_\xi}\), and the computation complexity is \(\bigO{n_\xi^2}\).

We explore the sparsity of the problem and decouple it into a two-stage linear program using the L-shaped algorithm \cite{birge2011introduction}. The algorithm returns the same optimal solution but reduces the computation cost from \(\bigO{n_\xi^2}\) approximately to \(\bigO{n_\xi}\) with a larger constant coefficient empirically.

The general structure of the algorithm is summarized in Algorithm \ref{alg:premission_lshaped}.
At the first stage, the algorithm solves a fixed-sized mixed-integer linear program (the size is not a function of the sample number \(n_\xi\)). At the second stage, the algorithm solves \(n_\xi\) small linear programs and then adds cuts to the first stage program according to the second-stage solutions. The two-stage process is iterated until convergence.

\begin{algorithm}[t]
\small

\textbf{Input:} Parameters of the optimization problem \eqref{eqn:bound_constraint}-\eqref{eqn:cvar_objective_h_sum_linear2}

\For{\(i \in M\)\textnormal{ and} \(a \in A\)}{
    \(L_{i a} = 0\)
}

\While{\textnormal{True}}{

    Solve the first stage problem 
    and let \((y_{1 i}^{[p]}, \ \cdots, \ y_{{n_v} i}^{[p]}, \ \lambda_{i a}^{[p]}, \ \theta_{i a}^{[p]}, x_{k i j}^{[p]})\) be the solution.

    \text{flag = True}

    \For{\(i \in M\)\textnormal{ and} \(a \in A\)}{
        {\color{ForestGreen}\textnormal{// \(L_{i a}\) is abbreviated to \(L\) in the following lines.}}
        
        \For{\(\xi = \)\textnormal{1 : }\(n_\xi\)}{
            Solve the second stage problem \eqref{eqn:premission_second_stage} and let \(\pi_{i a}^{(\xi) [p]}\) be the Lagrangian multiplier associated with the solution \(w_{i a}^{(\xi) [p]}\). 
        }
        
        Calculate \(D^{[L]}_{i a}\) and \(d^{[L]}_{i a}\) according to \eqref{eqn:premission_second_stage_constraint}.
    
        \If{\(D^{[L]}_{i a} [y_{1 i}^{[p]}, \ \cdots, \ y_{{n_v} i}^{[p]}, \ \lambda_{i a}^{[p]}]^\transpose + \theta_{i a}^{[p]} < d^{[L]}_{i a}\)}
        {
            \text{flag = False}
    
            \(L = L + 1\)
            
            Add the cut \(D^{[L]}_{i a} [y_{1 i}, \ \cdots, \ y_{{n_v} i}, \ \lambda_{i a}]^\transpose + \theta_{i a} \geq d^{[L]}_{i a}\)
        }
    }
    \If{\textnormal{flag}}
    {
        \Return the solution \(x_{kij}^{[p]}\) {\color{ForestGreen}\textnormal{// Optimality obtained.}}
    }
    \(p = p + 1\) {\color{ForestGreen}\textnormal{// The problem will contain more cuts \eqref{eqn:premission_first_stage4}.}}
}

\caption{L-shaped algorithm for the model.}
\label{alg:premission_lshaped}
\end{algorithm}

The first stage problem in the algorithm is
\begin{align}
    \min &\ C_e \underset{k \in V}{\sum} \ \underset{i,j \in \node}{\sum} b_{k i j} \cdot x_{k i j} + C_{q} \underset{i \in U}{\sum} q_i 
    + C_h \underset{i \in M}{\sum} h_i \tag{\ref{eqn:nonlin_penalty_objective}} \\ 
    \text{s.t.} & \text{ \eqref{eqn:bound_constraint}-\eqref{eqn:task_complete_constraint1} and} \nonumber \\ 
    h_{i} &= \sum_{a \in A \text{ s.t. } \gamma_{i a} \neq 0 } h_{i a}, \quad \quad \quad \quad \quad \quad \quad \quad \ \ \ \forall i \in M, \tag{\ref{eqn:nonlin_cvar_objective_h}} \\
    h_{i a} &\geq \eta_\beta \left({-c_{k a} + \gamma_{i a}} \right) \cdot r_{k i}, \ \ \forall k \in V, a \in A, \forall i \in M, \tag{\ref{eqn:lin_cvar_objective_h_min}} \\
    h_{i a} &= \lambda_{i a} + \frac{1}{n_\xi (1 - \beta)} \theta_{i a}, \quad \quad \quad \quad \ a \in A, \forall i \in M, \label{eqn:premission_first_stage3}\\
    D^{[\ell]}_{i a} & [y_{1 i}, \ \cdots, \ y_{{n_v} i}, \ \lambda_{i a}]^\transpose + \theta_{i a} \geq d^{[\ell]}_{i a}, \nonumber \\  \quad
    & \quad \quad \quad \quad \quad \quad \quad \forall \ell = 1,\cdots,L, \quad a \in A, \forall i \in M. \label{eqn:premission_first_stage4}
\end{align}

If capability \(a \in A\) is noncumulative, \(h_{i a}\) is calculated according to \eqref{eqn:lin_cvar_objective_h_min}. If capability \(a \in A\) is cumulative, the large linear program described in \eqref{eqn:cvar_objective_h_sum_linear2} is replaced with \eqref{eqn:premission_first_stage3}-\eqref{eqn:premission_first_stage4}. \(\theta_{i a}\) is a lower bound helper variable for \(\sum_{\xi = 1}^{n_\xi} w_{i a}^{(\xi)}\) in \eqref{eqn:cvar_objective_h_sum_linear2}, and is tightened iteratively with the cuts in \eqref{eqn:premission_first_stage4} obtained from the second stage.

In one iteration, as \(y_{k i}\) and \(\lambda_{i a}\) are determined in the first stage and fixed temporarily, the large linear program in \eqref{eqn:cvar_objective_h_sum_linear2} can be decoupled into \(n_\xi\) small linear programs with analytic solutions. We call it the second stage problem:
\begin{align}
    \min \ & w_{i a}^{(\xi)} \nonumber \\
    \text{s.t. } & w_{i a}^{(\xi)} \geq -\underset{k \in V}{\sum} c_{k a}^{(\xi)} \cdot y_{k i}^{[p]} + \gamma_{i a}^{(\xi)} - \lambda_{i a}^{[p]}, \label{eqn:premission_second_stage} \\
    & w_{i a}^{(\xi)} \geq 0. \nonumber
\end{align}

Once the second stage problem is solved during each iteration, an additional optimality cut can be added in \eqref{eqn:premission_first_stage4} according to the Lagrangian multipliers (simplex multipliers) \(\pi^{(\xi) [p]}_{i a}\) associated with the second stage solutions. The parameters of the new optimality cut \(D^{[L]}_{i a}\) and \(d^{[L]}_{i a}\), are calculated as follows. Note that \(L\) is an integer label for the new cut.
\begin{align}
\begin{split}
    D^{[L]}_{i a} &= \sum_{\xi = 1}^{n_\xi} \pi^{(\xi) [p]}_{i a} \left[ c_{1 a}, \ \cdots, \ c_{{n_v} a}, \ 1 \right]. \\
    d^{[L]}_{i a} &= \sum_{\xi = 1}^{n_\xi} \pi^{(\xi) [p]}_{i a} \cdot \gamma_{i a}^{(\xi)}.
\end{split}\label{eqn:premission_second_stage_constraint}
\end{align}

The Lagrangian multipliers \(\pi^{(\xi) [p]}_{i a}\) are obtained by solving the second stage dual problem \eqref{eqn:premission_second_stage_dual1}, and the solutions are shown in \eqref{eqn:pi_solution}.
\begin{align}
    \max \ & \left( -\underset{k \in V}{\sum} c_{k a}^{(\xi)} \cdot y_{k i}^{[p]} + \gamma_{i a}^{(\xi)} - \lambda_{i a}^{[p]} \right) \cdot \pi^{(\xi)}_{i a} \label{eqn:premission_second_stage_dual1} \\
    \text{s.t. } & \ 0 \leq \pi^{(\xi)}_{i a} \leq 1. \nonumber \\
     \pi^{(\xi) [p]}_{i a} = &
    \begin{cases}
    0, \ -\underset{k \in V}{\sum} c_{k a}^{(\xi)} \cdot y_{k i}^{[p]} + \gamma_{i a}^{(\xi)} - \lambda_{i a}^{[p]} < 0 \\
    1, \ \text{otherwise}
    \end{cases}. \label{eqn:pi_solution}
\end{align}

%% file: section/flow_cover.tex
\section{Decompose Agent Flows into Paths}\label{sec:flow_decompose}
Solving the teaming problem in Sec. \ref{sec:problem_model} using the algorithm in Sec. \ref{sec:lshaped_algorithm} provides the optimal solution to \(x_{k i j}, \forall k \in V, \forall i, j \in \node\), i.e., the flows of different agent species.
This section discusses an optimal algorithm to extract a routing plan for each individual agent from the agent flows.

The flow decomposition problem contains two steps: rounding and splitting. First, we round the agent network flow obtained from the MIP risk minimization model in the previous section. Secondly, we split the network flow into a set of agent routes to obtain a plan for each individual agent.

\begin{figure}[t]
    \centering
	\subfloat[\label{fig:graphical_model_continuous_flow}]{
    	\includegraphics[width=0.88\linewidth, trim=0 0 0 0, clip]{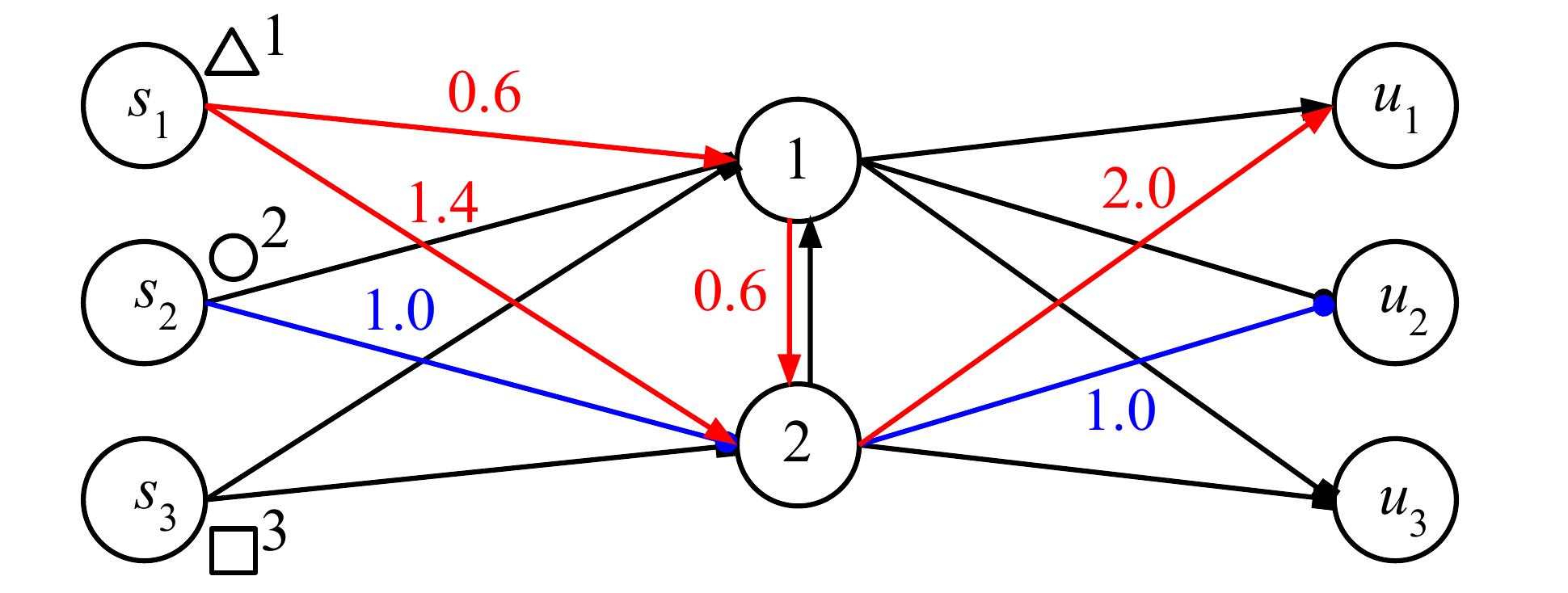}}
    \hfill
	\subfloat[\label{fig:graphical_model_continuous_v1}]{
    	\includegraphics[width=0.88\linewidth, trim=0 0 0 0, clip]{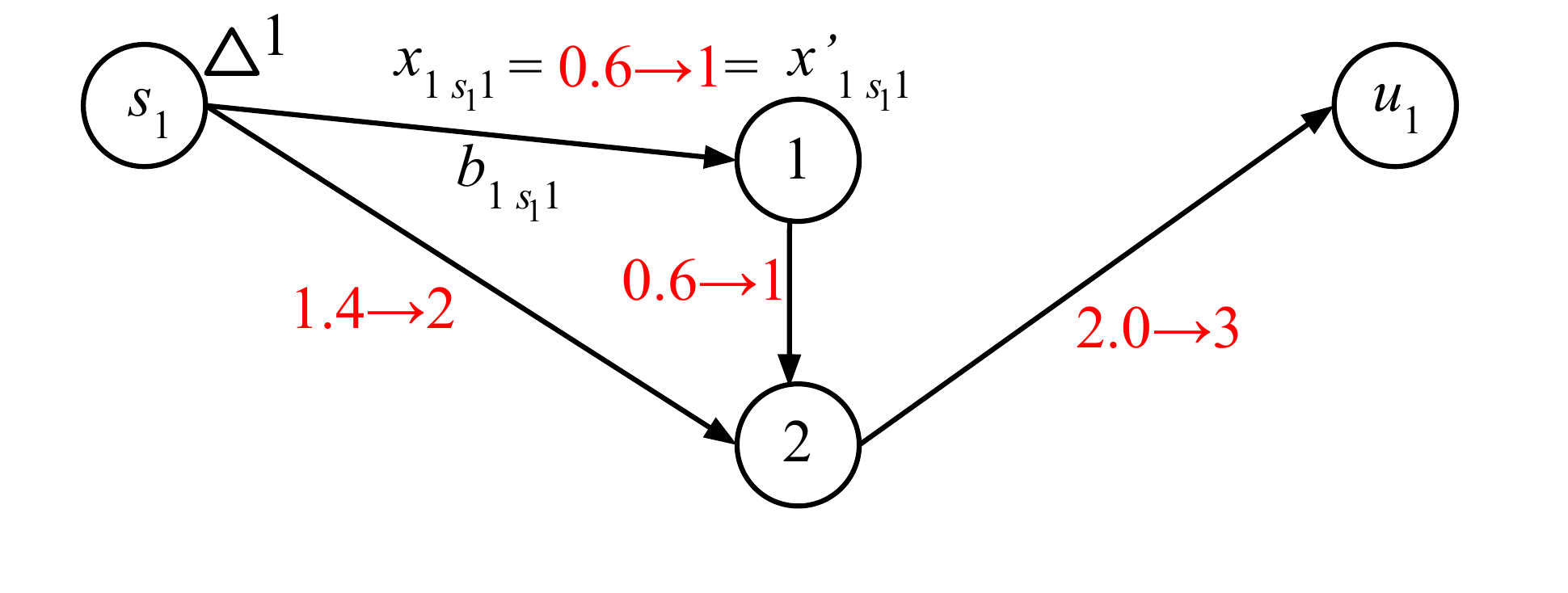}}
    \hfill
	\subfloat[\label{fig:graphical_model_continuous_v1_cover}]{
    	\includegraphics[width=0.88\linewidth, trim=0 0 0 0, clip]{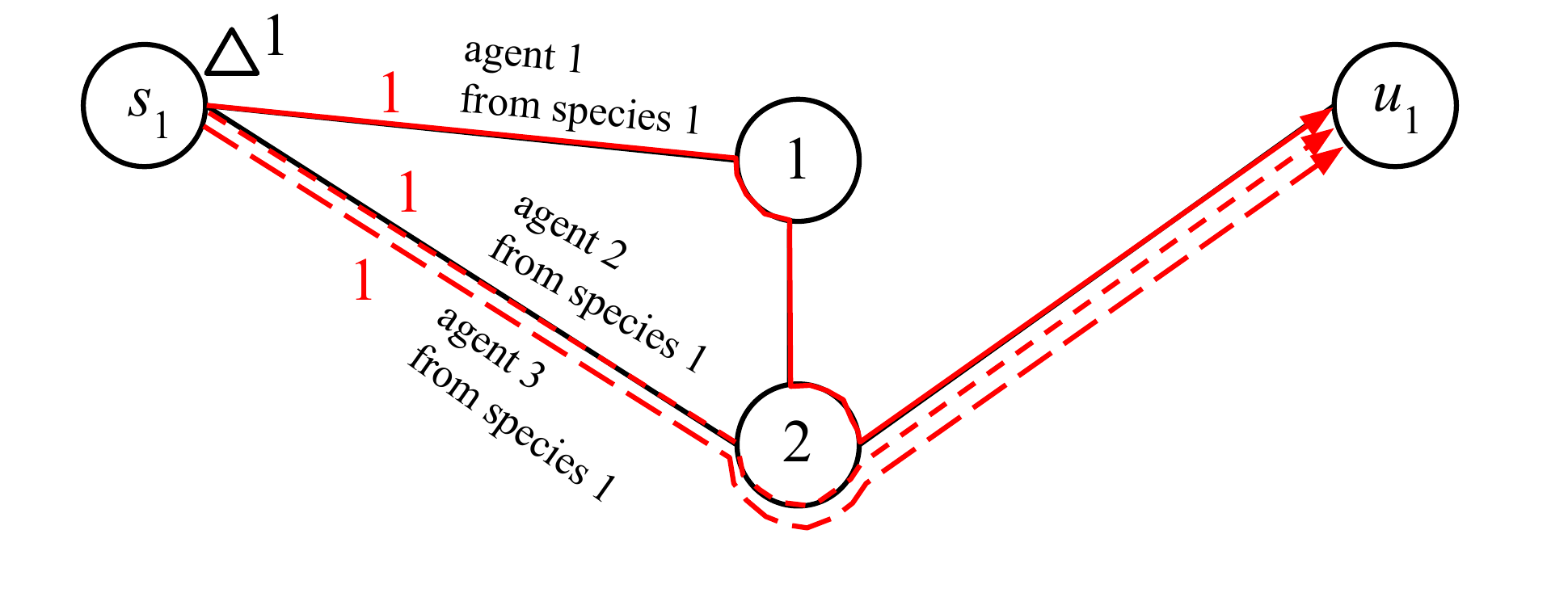}}
    \hfill
	\caption{Decompose a flow into paths. The \(x\)'s are the flow on an edge, with the red and blue numbers an example of the actual value. The colors distinguish the agent species. The \(b\)'s are the energy cost of the edge. (a) Resulted agent flows. (b) The flow of agent species \(1\). (c) The split of agent species \(1\) using three individuals.}
	\label{fig:decompose_flow}
\end{figure}

An example is shown in Fig. \ref{fig:decompose_flow}. The flow network of species \(1\) is extracted and rounded off in Fig. \ref{fig:graphical_model_continuous_v1}. Then, in Fig. \ref{fig:graphical_model_continuous_v1_cover}, three individual agent paths with flows of 1 on the path edges are selected such that the sum of the flows for the three paths equals the rounded integer network flow in Fig. \ref{fig:graphical_model_continuous_v1}. Note that in order to maintain the task requirements, we have to round up a fractional flow rather than round it to the nearest integer.

In short, requirements for the rounding problem include:
\begin{itemize}
    \item Round up the agent flow to integers.
    \item Maintain the flow constraint. (E.g., if two agents enter a task, they should also leave the task.)
    \item The rounding should result in minimum additional energy cost. (There can be multiple solutions with only the first two constraints.)
\end{itemize}

Requirements for the splitting problem include:
\begin{itemize}
    \item The integer agent flow should be split into multiple individual routes.
    \item The most costly routes due to the current split should be minimized. (There can be multiple solutions with only the first constraint.)
\end{itemize}

There is no trivial solution to the rounding and splitting problems. Additional information and examples are given in Appendix \ref{sec:appendix_flow}. In the following two sections, we formalize two optimization problems to determine the optimal flow decomposition.

\subsection{Minimum Energy Cost Rounding}\label{sec:flow_round_up}

Formally, a flow network is a graph \(G(N = S \cup M \cup U, E)\) with a function \(f(\cdot): E \rightarrow \mathbb{R}\), where \(S\), \(M\), \(U\), and \(E\) denote the set of start nodes, intermediate nodes, terminal nodes, and directed edges, respectively. For a node \(m \in M\), suppose the incoming and outgoing edge sets are \(E_{m}^{\text{in}}\) and \(E_{m}^{\text{out}}\), respectively. Then, the function \(f(\cdot)\) should satisfy the flow constraint
\begin{align}
    \sum_{e \in E_{m}^{\text{in}}} f(e) = \sum_{e \in E_{m}^{\text{out}}} f(e), \quad \forall m \in M. \nonumber 
\end{align}

In this section, the flow function \(f_k(i,j) = x_{kij}\) indicates the number of agent species \(k \in V\) on edge \((i, j) \in E\).

See Fig. \ref{fig:graphical_model_continuous_v1} as an example. Suppose the flow output from the MIP model in Sec. \ref{sec:problem_model} for agent species \(k \in V\) is \(x_{k i j}\) on edge \((i, j)\). Let the integer flow after the rounding process be \(x_{k i j}'\). Then, the linear program in below will return an integer flow network with minimum energy cost.
\begin{align}
    \min & \sum_{i, j \in \node} b_{k i j} \cdot x_{k i j}' & \label{eqn:round_obj}\\
    \text{s.t. } & x_{k i j}' \geq {\lceil x_{k i j} \rceil}, & \forall x_{kij} > 0, \nonumber \\
    & x_{k i j}' = 0,  & \forall x_{kij} = 0. \label{eqn:round_bound} \\
    & \sum_{i \in S \cup M} x_{k i m} = \sum_{j \in U \cup M} x_{k m j}, & \forall m \in M. \label{eqn:round_flow_constraint}
\end{align}

The objective function \eqref{eqn:round_obj} penalizes the energy cost. \eqref{eqn:round_bound} ensures the rounding is happening upward, where \(\lceil x_{k i j} \rceil\) denotes the smallest integer larger than \(x_{k i j}\). The network flow constraint \eqref{eqn:round_flow_constraint} should be maintained during the optimization. \(S\), \(U\), and \(M\) are the set of start, terminal, and task nodes, respectively.

Note that there is no explicit integer constraint to ensure that \(x_{kij}'\) is an integer. However, if this linear program is solved using Simplex-based algorithms, the solutions to \(x_{kij}'\) are guaranteed to be integers. The proof is given in Appendix \ref{sec:appendix_integer}.
Because of this, the minimum energy rounding problem could be solved in polynomial time through a linear program (instead of an integer linear program).

\subsection{Minimum Max-Energy Flow Split}\label{sec:flow_minmax_cover}

\begin{figure}[h!]
	\centering
	\subfloat[]{
    	\includegraphics[width=0.9\linewidth, trim=0 0 0 0, clip]{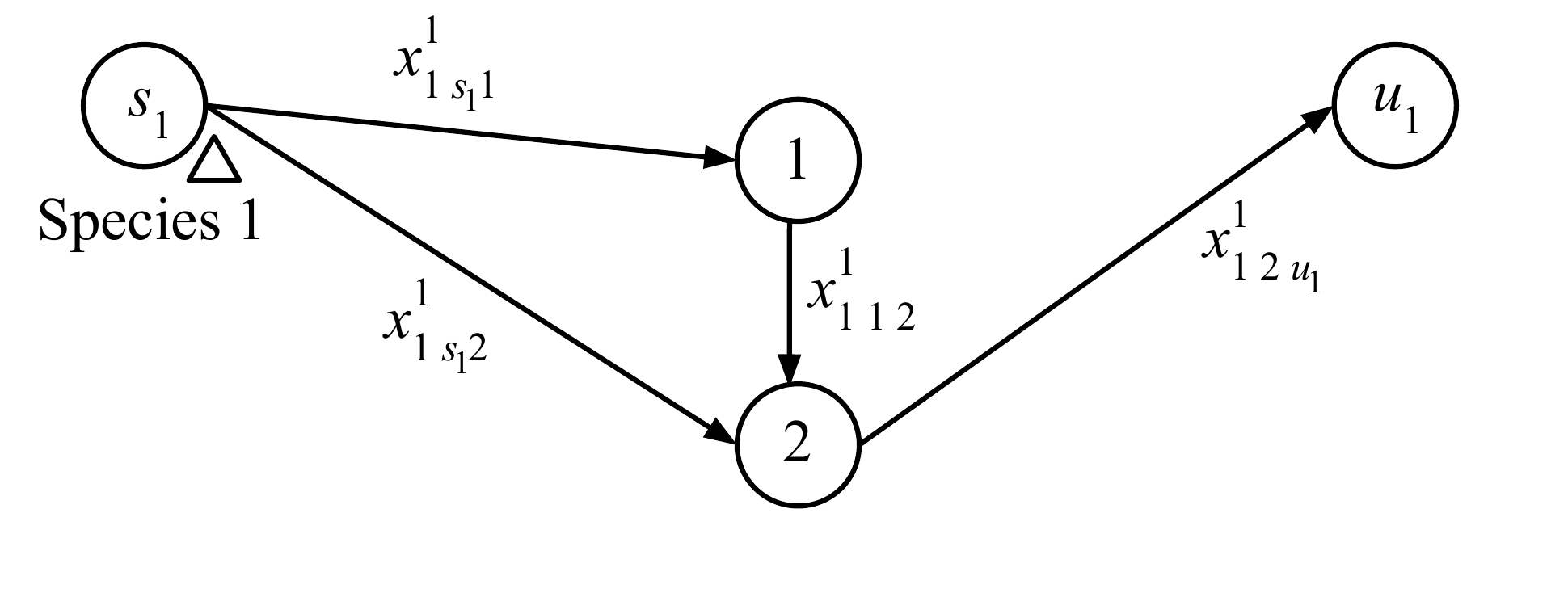}}
    \hfill
	\subfloat[]{
    	\includegraphics[width=0.9\linewidth, trim=0 0 0 0, clip]{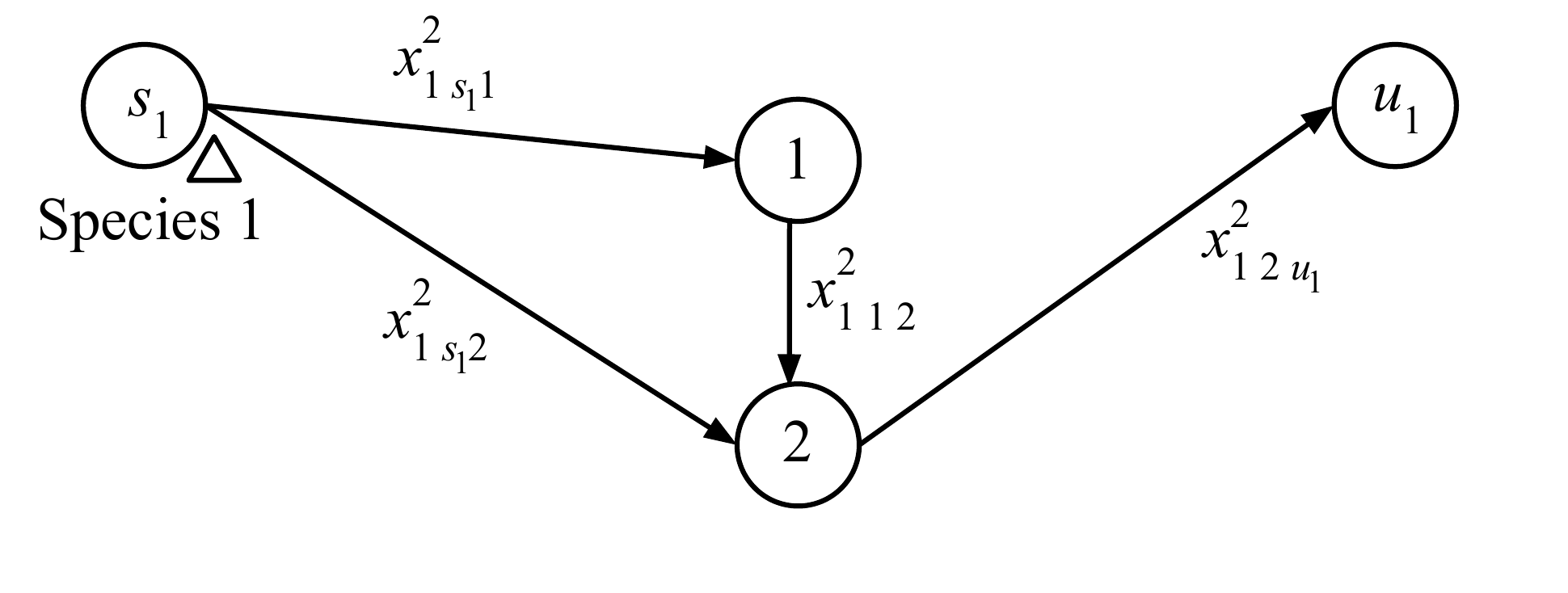}}
    \hfill
	\subfloat[]{
    	\includegraphics[width=0.9\linewidth, trim=0 0 0 0, clip]{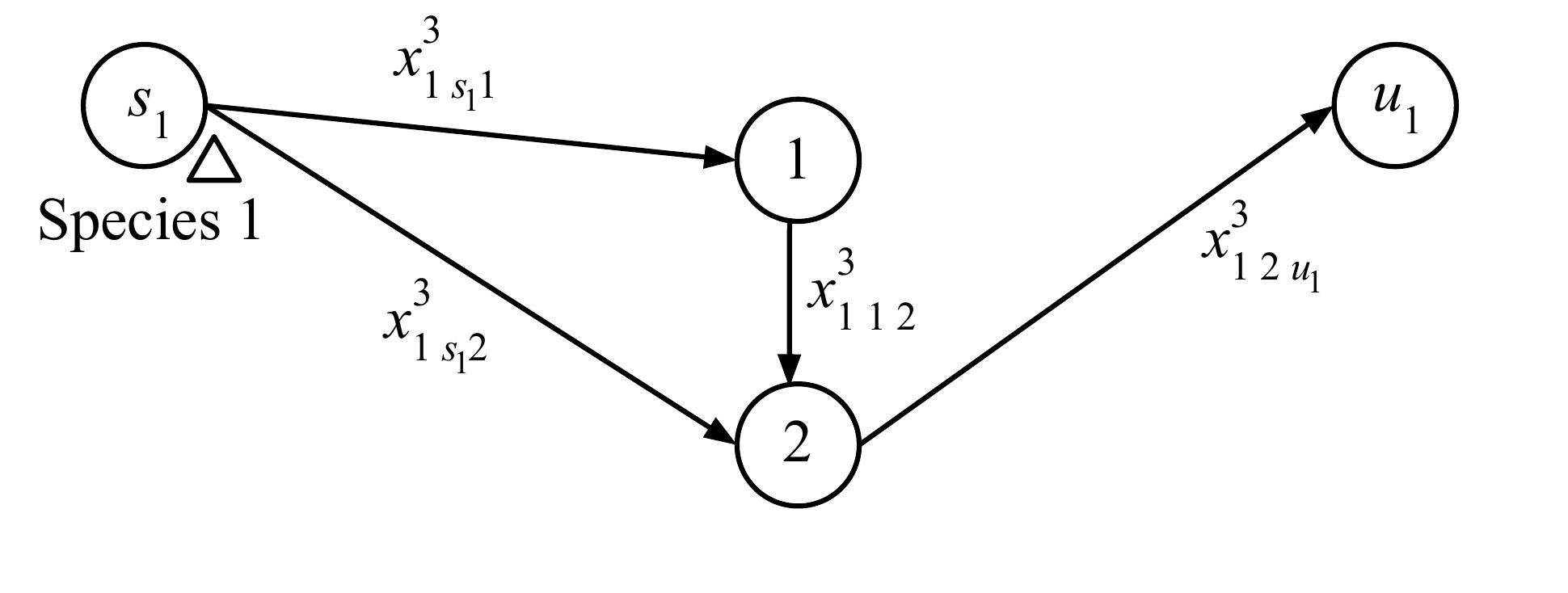}}
    \hfill
	\caption{Graphical model for the minimum max-energy flow split problem for agent species \(1\). The \(x\)'s are the flow on an edge. (a)(b)(c) are the flow networks for the three agents of species \(1\).}
	\label{fig:flow_cover_problem}
\end{figure}

After rounding, we obtain the integer flows \(x_{k i j}'\) in Fig. \ref{fig:graphical_model_continuous_v1}, the next step is to split this integer flow into individual agent paths in Fig. \ref{fig:graphical_model_continuous_v1_cover}. By summing the out-going flow at the start node, we are able to get the needed number of agents from species \(k \in V\), denoted as \(n'_k\). Then we compose \(n'_k\) graphs as in Fig. \ref{fig:flow_cover_problem} for the \(n'_k\) agents.

We can formalize an integer linear program (ILP) as follows; find the routes for all agents, such that the maximum individual energy cost is minimized and the unit agent flow on the paths sums to the original network flow.
\begin{align}
    & \quad \min_{x_{k i j}^{l}} \ \max_{l} \sum_{i, j \in \node} b_{k i j} \cdot x_{k i j}^{l} \label{eqn:path_convert_obj}\\
    & \text{s.t. } \sum_{l=1}^{n'_k} x_{k i j}^{l} = x_{k i j}', \quad \forall i,j \in \node, \label{eqn:path_convert_sum_flow}\\
    & \sum_{i \in \node} x_{k i m}^l = \sum_{j \in \node} x_{k m j}^l, \quad \forall m \in M, \ \ \forall l = 1,\cdots,n'_k, \label{eqn:path_convert_flow_constraint} \\
    & \quad \ \ \sum_{i \in M} x_{k s i}^l = 1, \quad \forall l = 1,\cdots,n'_k, \label{eqn:path_convert_flow_constraint2} \\
    & \quad \quad x_{k i j}^l  \in \{0, 1\}, \quad \forall i,j \in \node, \quad \forall l = 1,\cdots,n'_k. \nonumber
\end{align}

In the ILP above, the objective \eqref{eqn:path_convert_obj} penalizes the maximum energy cost of an individual agent. Equation \eqref{eqn:path_convert_sum_flow} ensures that the resulting agent flows in sum up to the rounded flow. Equation \eqref{eqn:path_convert_flow_constraint} is the network constraint. Equation \eqref{eqn:path_convert_flow_constraint2} requires that in each subgraph, there is only one agent.

%% file: section/experiment_result.tex
\section{Experiments and Results}\label{sec:experiments}

In this section, we first initialize randomized flow networks to test the decomposition of the agent flows, and then use two practical applications to evaluate the robustness, generalizability, and scalability of the proposed risk minimization model. The models and algorithms are implemented using the GUROBI solver. All computations were done on a laptop with Intel i7-7660U CPU (2.50GHz).

\subsection{Decompose Agent Flows into Paths}
In this section, we use randomized test cases of different sizes to evaluate the computational cost of the flow decomposition process. The evaluation metric is the time to solve both the rounding and splitting problems to zero optimality gaps.

\subsubsection{Setup and Test Cases}
An agent flow network with random connections is initialized, and the flow and unit cost of each edge is sampled from uniform random distributions. The hyper-parameters are the maximum flow on an edge and the node number of the network. The optimization models in Sec. \ref{sec:flow_decompose} are then applied to the initialized random flow network to solve the minimum energy cost rounding and minimum max-energy flow split problems.

\subsubsection{Result and Discussion}

Different sizes of flow decomposition problems are solved, and the computational costs of the two steps are listed in TABLE \ref{tab:flow_decompose_result}. The size of the rounding and split problems are proportional to edge number and edge number \(\times\) sum integer flow, respectively. For the cases shown in TABLE \ref{tab:flow_decompose_result}, the rounding steps can be completed within several milliseconds. The rounding process scales well because the linear program can be solved in polynomial time. The split step involves solving an integer linear program, which does not scale well generally with the problem size. The largest test cases shown in the table involve 70 tasks and 241 agents on average, much larger than the typical teaming problem sizes that the risk minimization model will be applied to. Therefore, the flow decomposition part will not be the bottleneck of the overall teaming planner.

\begin{figure}[hbt!]
	\centering
	\includegraphics[width=0.8\linewidth]{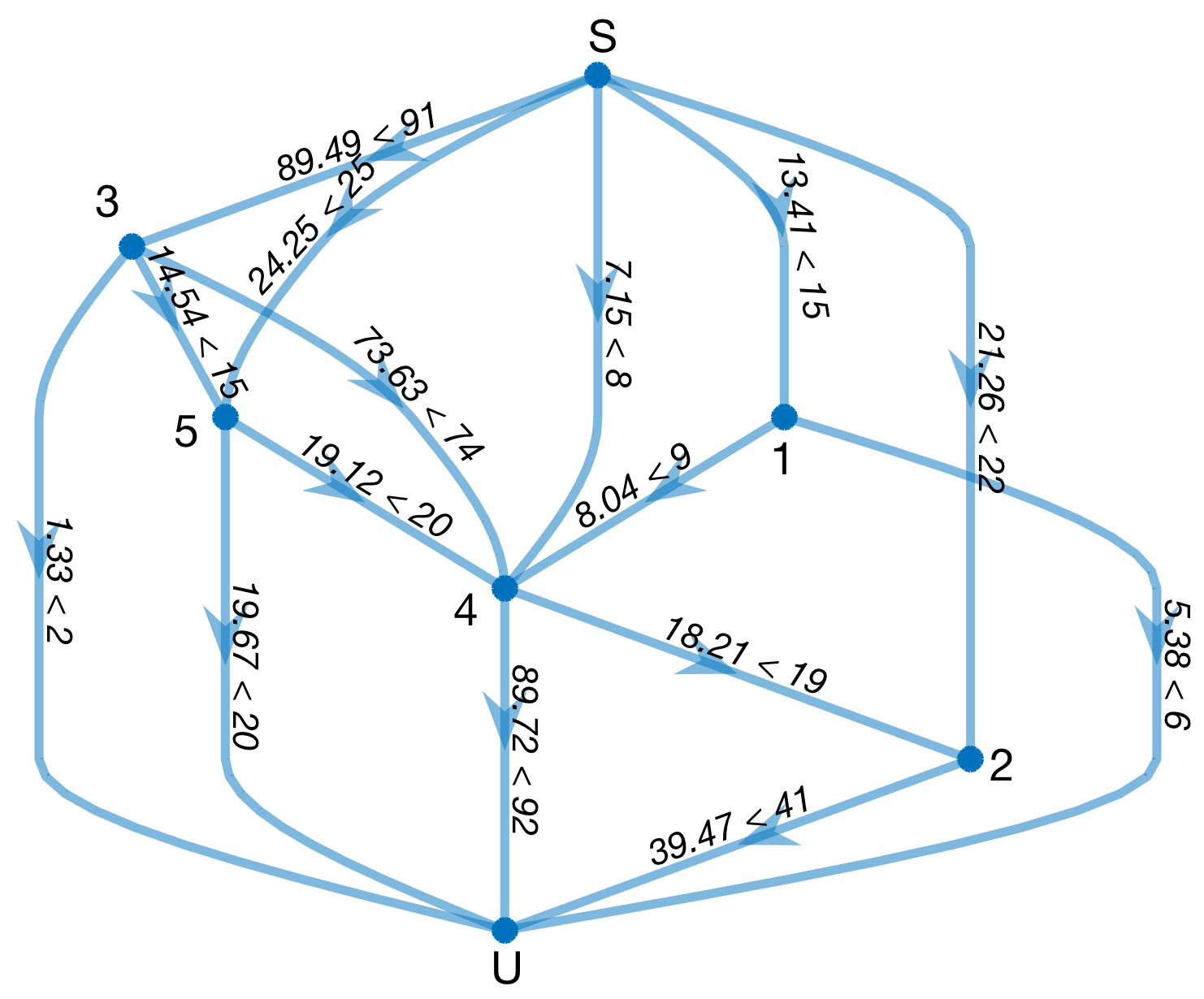}
	\caption{Flow rounding results (5 task nodes, 15 edges). There are two numbers on each edge, indicating the agent flows before and after the rounding process. This example corresponds to one instance in the first row of TABLE \ref{tab:flow_decompose_result}.}
	\label{fig:flow_round_result}
\end{figure}

Though there is no explicit integer constraint in the rounding model in Sec. \ref{sec:flow_round_up}, we provided a proof in Appendix \ref{sec:appendix_integer} that showed that the solutions would be integers, and the results support it. As an example, the rounding solution for a case from the first row of the table is shown in Fig. \ref{fig:flow_round_result}, where we can see that the agent flow on each edge is rounded up to an integer, while the overall network flow constraint is maintained. For instance, the 13.41 on edge (S, 1) is rounded to 15 instead of 14 in order to maintain the flow constraint.

\newcommand{\stdsize}[1]{\scriptsize \(\pm\)#1}

\begin{table}[t]
  \caption{The computational cost of the flow decomposition. }
  \label{tab:flow_decompose_result}%
    \begin{tabular}{cccccc}
    \toprule
    Task\(^*\)  & Edge  & Flow  & Int flow & Round time & Split time \\
    \midrule
    5     & 11\stdsize{1}    & 110.6\stdsize{33.7} & 114\stdsize{34}   & 0.001\stdsize{0.0005} & 0.09\stdsize{0.04}   \\
    10    & 28\stdsize{3}    & 21.3\stdsize{2.8}   & 30\stdsize{2}     & 0.001\stdsize{0.0004} & 0.13\stdsize{0.09}   \\
    10    & 26\stdsize{3}    & 213.7\stdsize{30.5} & 221\stdsize{31}   & 0.001\stdsize{0.0002} & 1.23\stdsize{0.89}   \\
    35    & 107\stdsize{8}   & 135.7\stdsize{13.2} & 172\stdsize{13}   & 0.001\stdsize{0.0001} & 19.33\stdsize{9.50}  \\
    70    & 210\stdsize{7}   & 163.6\stdsize{7.5}  & 241\stdsize{8}    & 0.001\stdsize{0.0009} & 197.0\stdsize{32.9}  \\
    \bottomrule
    \end{tabular}%
    \\
    \\
    * `Task' is the number of tasks, which equals the number of nodes in the flow network. `Flow' is the sum of agent flows from the start to the terminal node and `int flow' is the value after rounding (i.e., the number of agents used). `Round time' and `split time' are the times used to solve the rounding and splitting problems, with units in seconds. For each row, we randomly initialize 10 similar sized test cases to evaluate the algorithm and show the mean and standard deviation.
\end{table}%

\subsection{Capture the Flag}\label{sec:capture_flag}
\begin{figure}[hbt!]
	\centering
	\includegraphics[width=1\linewidth]{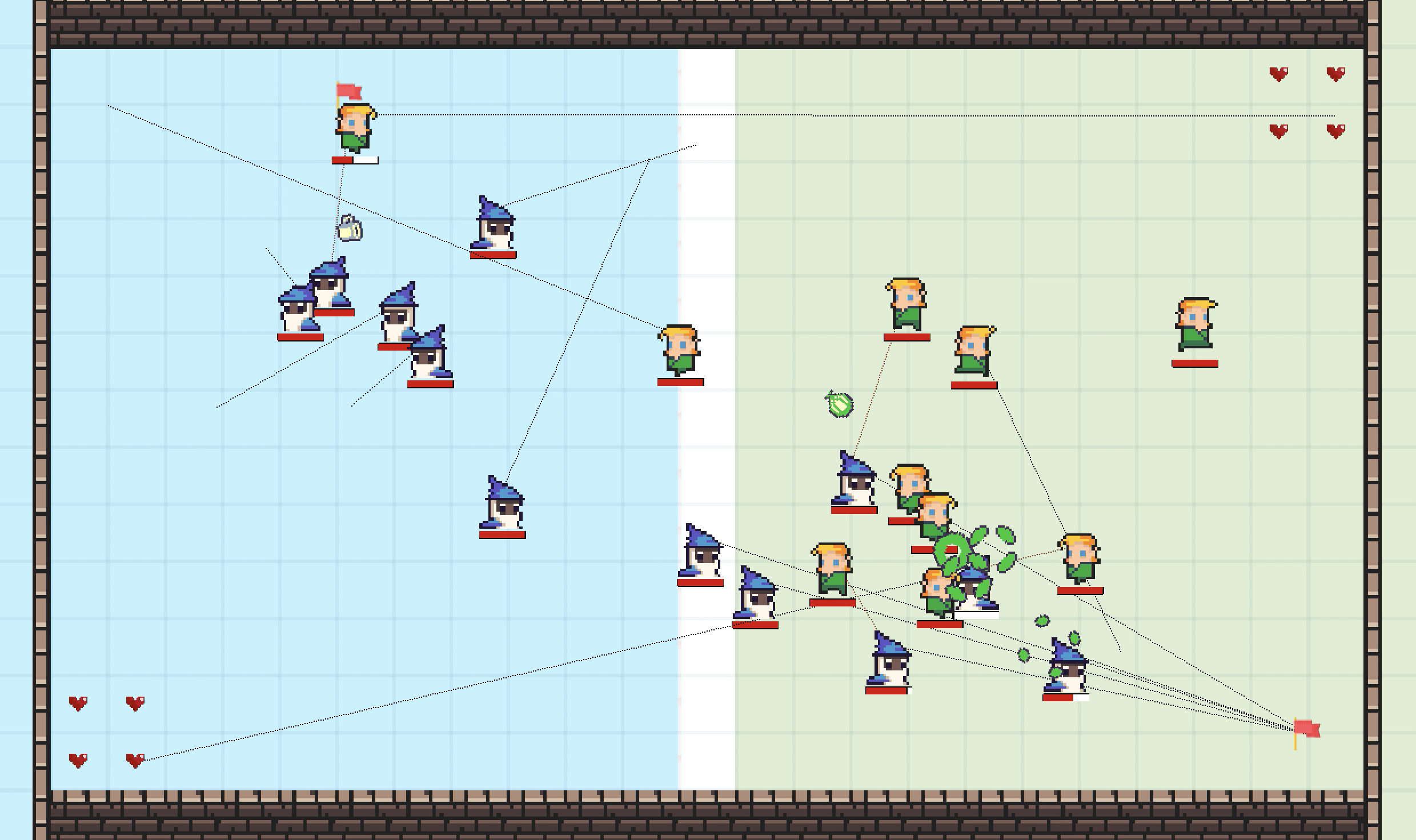}
	\caption{Capture the flag setup. Each small square tile is \(1 \times 1\) unit length. A line points to the current target of the corresponding agent.}
	\label{fig:capture_flag_setup}
\end{figure}

In this section, we apply the risk minimization model in Sec. \ref{sec:problem_model} to a team of agents in a capture the flag game setting and compare the team performance against the baseline (STRATA \cite{ravichandar2020strata}) in a simulation environment shown in Fig. \ref{fig:capture_flag_setup}.
The goal of this simulation is to evaluate and demonstrate the performance of the task assignment component of our framework. The number of wins is used as the metric for task performance.

\subsubsection{Game Setup, Baseline, and Metrics}

The blue and green teams contain 12 heterogeneous agents initialized at random locations within their own sides. The overall goal is to win the game by capturing the flag from the other team. The 12 agents are from 4 species (3 individuals for each species). These species have different speed, viewing distance, health, and ammunition capabilities (TABLE \ref{tab:specie_capabililty}).
The specific values in the table are set according to the baseline paper \cite{ravichandar2020strata}. Among these capabilities, the first two are modeled as noncumulative, while the last two are cumulative.

Each agent is designed to play one role out of the 3 tasks listed in TABLE \ref{tab:task_description} and will disappear when its health reaches zero. The adversarial elements of the game are not explicitly modeled, and the focus is to assign tasks such that the requirements are fulfilled while overall time and energy costs are considered in the trade-off. Therefore, the agents follow predefined behaviors once their tasks are determined.

STRATA \cite{ravichandar2020strata} is chosen as the baseline for this work because its model makes use of similar concepts like agents, species, capability vectors, and task requirements specified by capabilities.
Since there is no scheduling involved in this game (the tasks are all `life-long' tasks), \modelrisk{} reduces to an instantaneous task assignment, tackling the same type of problem as STRATA.
Therefore, this game will provide a comparison between STRATA and the task assignment component of \modelrisk{}. 
In addition to STRATA, we also compare our model to a random task assignment mechanism where an agent is randomly assigned a task.

All games are played with two teams using different assignment models. The metric used in this simulation is the number of wins in 500 games, which reveals the relative task assignment performance. A win is either a team taking back the other side's flag or defeating all enemies through a fight mode. A draw happens when all of the agents from both sides decide to defend and no longer attack, or if no team wins the game within 120 seconds.

Both STRATA and our model need task requirements specified as capability distributions. For the three tasks, the agents are programmed to switch to a healing mode automatically when their health is lower than a threshold and return to attack/defend after they heal. 
Namely, the healing role is a temporary state that happens when we think an agent is too `unhealthy' to continue. In the game, \modelrisk{} and the baseline assign either the attack or defend role to the agents.
The threshold of entering the healing state is set to 33\% of an agent's full health. We have done a parametric study, and results show that the threshold does not affect the number of wins and the task assignment patterns that will be discussed.
The required capabilities of the attack and defend tasks are listed in TABLE \ref{tab:task_requirement}. The task requirements in this game are deterministic. However, since the agent capabilities in Table \ref{tab:specie_capabililty} are stochastic, the planning problem is still stochastic.

The specific numbers are manually decided to reflect the following requirements: the attack task needs agents with higher speed and health to capture the flag, while the defend task needs agents with higher viewing distance and ammunition to detect and defeat the opposing team. In TABLE \ref{tab:task_requirement}, 1131 is 65\% of the expected total initial health of the 12 agents, and 231 is 55\% of the expected total ammunition.

\begin{table}[htb!]
  \begin{center}
  \caption{Capability distributions of the four agent species.}
  \label{tab:specie_capabililty}%
    \begin{tabular}{c|cccc|cccc}
    \toprule
    species & \({\mu_1} ^*\)     & \(\mu_2\)     & \(\mu_3\)     & \(\mu_4\)     & \(\sigma_1^2\)     & \(\sigma_2^2\)     & \(\sigma_3^2\)     & \(\sigma_4^2\) \\
    \midrule
    1     & 1.5   & 2     & 90    & 40    & 0.35  & 0.1   & 10    & 3 \\
    2     & 1.5   & 4     & 60    & 40    & 0.35  & 0.1   & 10    & 3 \\
    3     & 3     & 2     & 80    & 30    & 0.35  & 0.1   & 10    & 3 \\
    4     & 3     & 4     & 350   & 30    & 0.35  & 0.1   & 10    & 3 \\
    \bottomrule
    \end{tabular}%
  \end{center}
    * \(\mu_i\) and \(\sigma_i^2\) (\(i =\) 1: speed, 2: view, 3: health, 4: ammunition) are the means and variances of the capability distributions (Gaussian in the experiment).
\end{table}%

\begin{table}[htb!]
  \centering
  \caption{General task descriptions.}
  \label{tab:task_description}%
    \begin{tabular}{c|l}
    \toprule
    Task & Description \\
    \midrule
    Attack   & Try to capture the flag of the other side. \\
    Defend   & Apply a defense mode to defeat opposing team members. \\
    Heal     & Use hearts to apply healing powers. \\
    \bottomrule
    \end{tabular}%
\end{table}%

\begin{table}[htb!]
  \caption{Task requirements are specified as capability distributions. }
  \label{tab:task_requirement}%
  \centering
    \begin{tabular}{c|cccc}
    \toprule
          & Speed & View  & Health & Ammunition \\
    \midrule
    Attack & \(\geq 2\)     &       & \(\geq 1131 \ (65\%)\) &  \\
    Defend &       & \(\geq 1\)     &       & \(\geq 231 \ (55\%)\) \\
    \bottomrule
    \end{tabular}%
\end{table}%

\subsubsection{Results}

Based on all the settings described in the above section, we developed a simulation environment in C++. A screenshot is shown in Fig. \ref{fig:capture_flag_setup}. The three task assignment models, random, baseline, and \modelrisk{}, are linked with the simulation environment. During the simulation, the mean time for the baseline and \modelrisk{} to optimize and output an assignment are 0.43 and 0.04 seconds, respectively.

\begin{figure}[hbt!]
	\centering
	\includegraphics[width=1\linewidth, trim=120 340 70 310, clip]{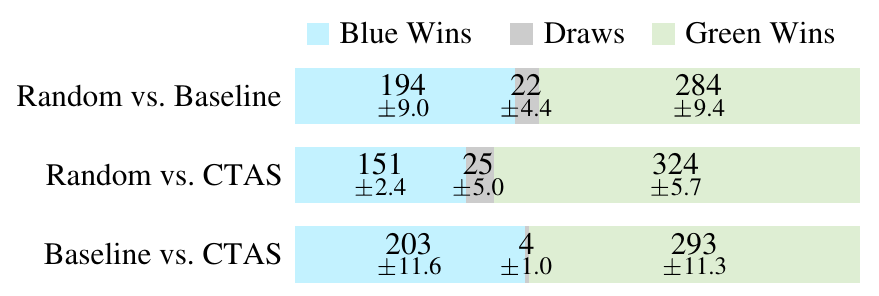}
	\caption{Relative performances of random task assignment, baseline, and \modelrisk{}. Each row shows the mean and standard deviation from ten 500 capture the flag games. The length of the bars are proportional to the mean. }
	\label{fig:capture_flag_win}
\end{figure}

\begin{figure}[h]
	\centering
	\includegraphics[width=0.9\linewidth, trim=80 245 70 260, clip]{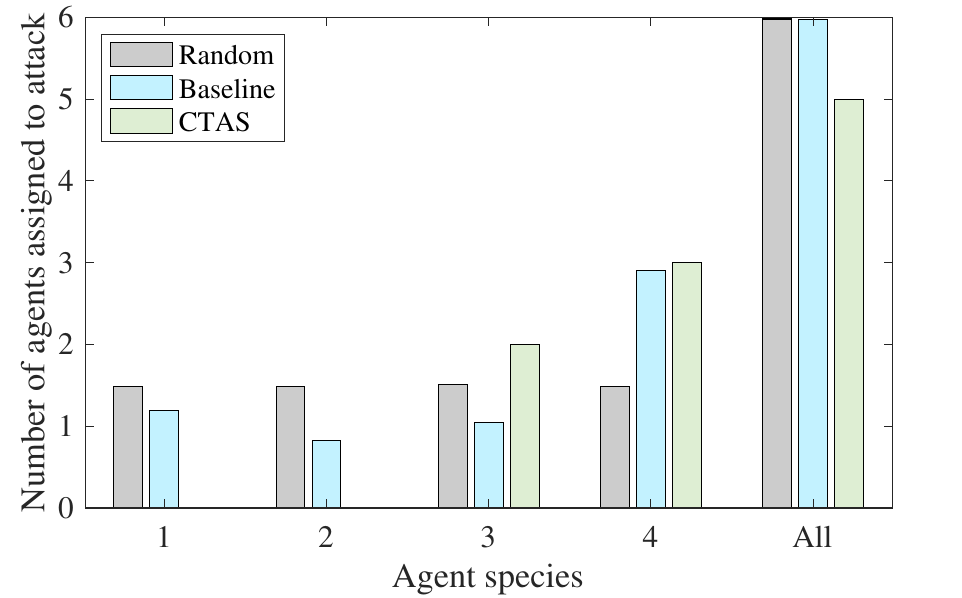}
	\caption{The average number of agents assigned to attack. \modelrisk{} is more likely to use species 4 and does not use species 1 and 2. The baseline shows a similar trend of preferring species 4 over the other species. This preference is logical since species 4 contains attributes of high speed and health, which are more suitable for the attack task than other species' attribute distributions (particularly species 1 and 2).}
	\label{fig:capture_flag_attack}
\end{figure}

The relative performances of the models are shown in Fig. \ref{fig:capture_flag_win}. 
As can be seen from Fig. \ref{fig:capture_flag_win}, \modelrisk{} results in a higher win rate as compared to the baseline or random selection methods. 
The average number of agents assigned to attack, grouped by species, is shown in Fig. \ref{fig:capture_flag_attack}.
Both the baseline and \modelrisk{} prefer using species 4 for the attacking task with the difference being that the baseline still uses species 1 and 2 for attack occasionally.

Though the baseline claims to be able to consider noncumulative capabilities through thresholding, it lacks an explicit mechanism to prevent incompetent agents from joining a task. For instance, though species 1 and 2 are not competent for the attack task due to their low speed, they are still allowed to conduct the task and contribute to other required capabilities such as health. However, this is not a good choice as there exist other agents that satisfy both the speed and health requirements of the attack task.

Another possible factor contributing to \modelrisk{}'s improved performance is that the \modelrisk{} algorithm directly minimizes the CVaR metric, which ensures enough task-required capabilities, whereas the baseline focuses on matching the expected requirements and penalizing variance of the assigned capability distributions.

In conclusion, through the comparison, our framework demonstrates superior task assignment performance against the baseline algorithm, and the task assignment patterns in Fig. \ref{fig:capture_flag_attack} support this performance result.

%% file: section/experiment_medical.tex
\subsection{Robotic Services during a Pandemic}\label{sec:explore_experiment}

In this section, we demonstrate the generality and the task assignment and scheduling components of our framework by considering a more complex scenario inspired by the COVID-19 pandemic.
We use our previous work in \cite{fu2020heterogeneous} as a baseline.
The energy cost and probability of success will be used to evaluate the optimality of the generated plans. The optimality gap given limited time for the optimization will be used to evaluate the scalability of the framework.

\subsubsection{Experiment Setup and Model Description}

Consider pandemic robotic services in a city environment consisting of multiple delivery, disinfection, test, and treatment sub-tasks that require one or multiple agents to complete.
The task-required capabilities and agent capabilities are modeled as random distributions.
We choose to include 8 types of tasks and 7 agent species in our investigation, some of the agents needing a human operator.

Applying the \modelrisk{} model to describe the problem of the pandemic robotic services, we first define 9 capability types according to the chosen tasks and agents in TABLE \ref{tab:capability_type}.

The 7 agent species and their capabilities are defined in TABLE \ref{tab:medical_vehicle_capability}, according to real-world references \cite{barfoot2020making}. The columns are different capabilities. The agent capabilities are Gaussian distributed random variables in this example. The values in the table are expectations of the random distributions, and the standard deviations are 10\% of the expectations.

\begin{figure*}[t]
	\subfloat[\label{fig:medical_task_distribution}]{
    	\includegraphics[width=0.53\linewidth, trim=90 280 60 260, clip]{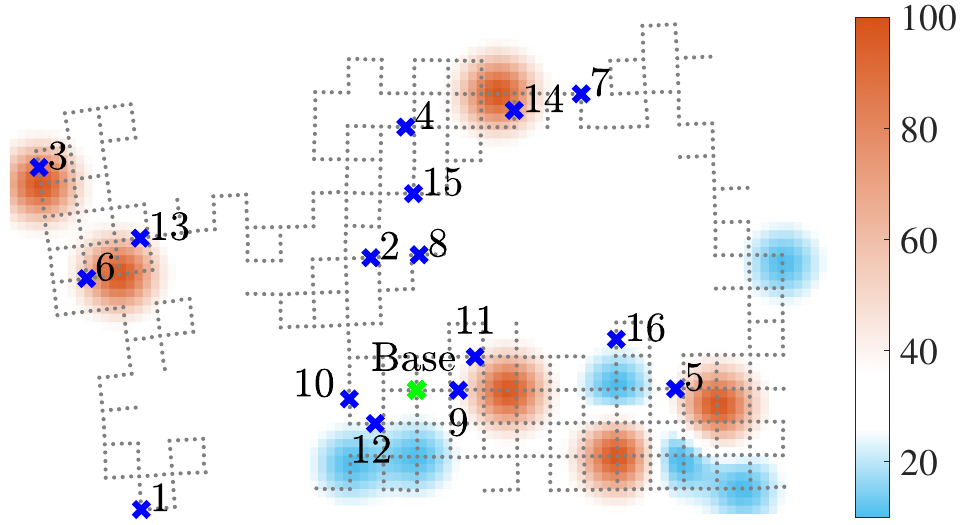}}
	\subfloat[]{
    	\includegraphics[width=0.4488\linewidth, trim=90 280 130 260, clip]{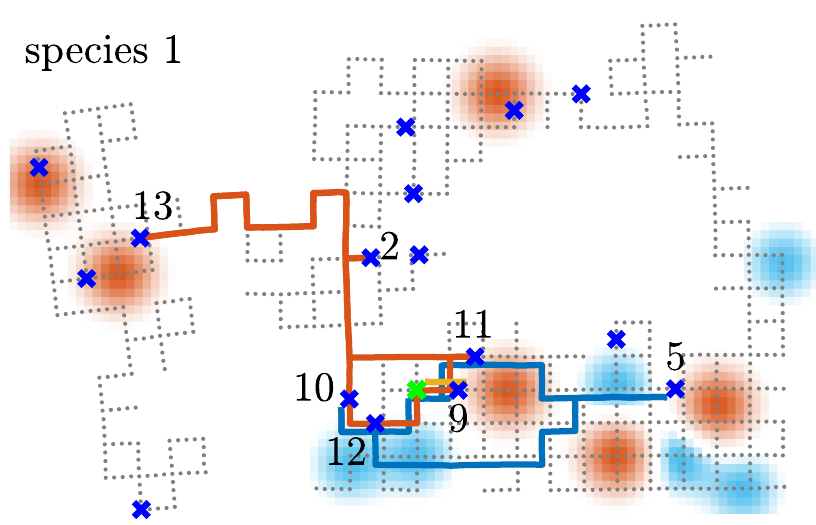}}
	\hfill
	\subfloat[]{
    	\includegraphics[width=0.4488\linewidth, trim=90 280 130 260, clip]{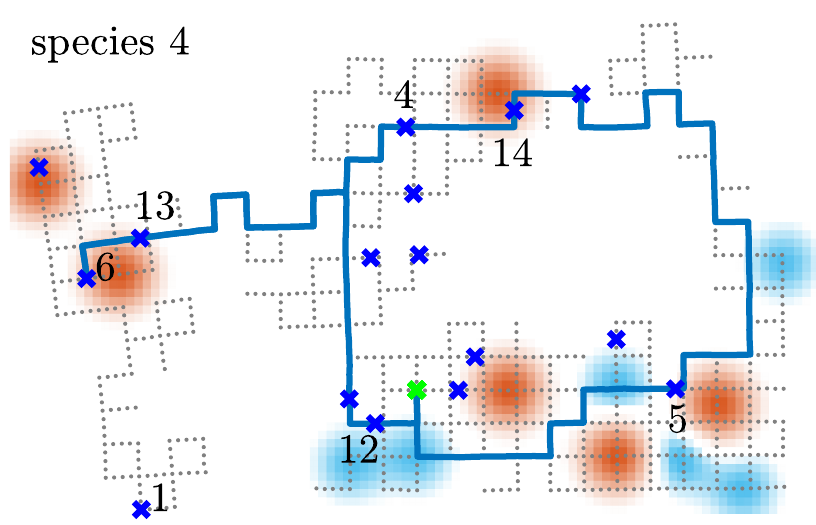}}
    \hspace{0.073\linewidth}
	\subfloat[]{
    	\includegraphics[width=0.4488\linewidth, trim=88 280 130 260, clip]{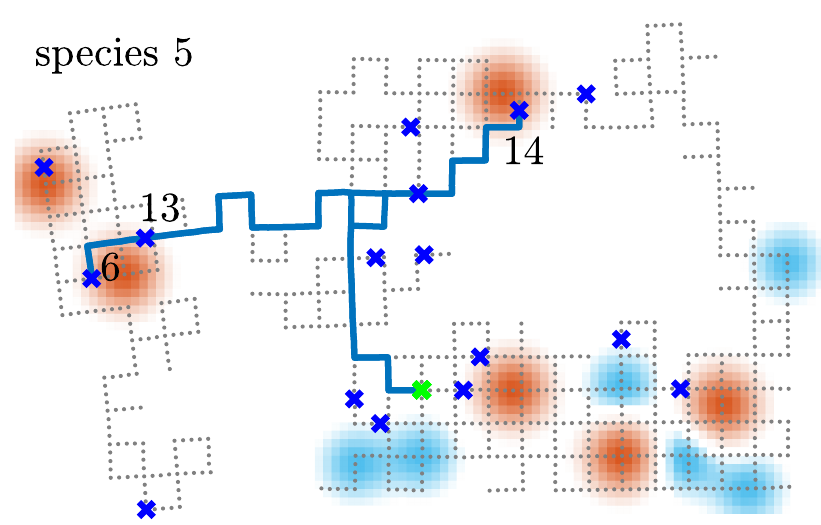}}
	\caption{(a) Task distribution: the 16 tasks are distributed in a city \cite{olson2006fast} where the unit travel cost depends on viral exposure. Blue and red stand for low and high energy costs, respectively. (b)-(d) The planned routes from the \modelrisk{} model for species \(1\), \(4\), and \(5\). Different colored lines represent distinct agent individuals from the same species.}
	\label{fig:medical_traj}
\end{figure*}

The 8 task types are described in TABLE \ref{tab:medical_task_description}. Their required capabilities are listed in TABLE \ref{tab:medical_task_requirement}.
The columns are different capabilities. A value in the table is the expected requirement for that specific capability, and the standard deviation is 10\% of the expectation. \(\gamma\) is a scaling coefficient. E.g., the \(\gamma\) in (row 1, column 3) means task type \(1\) requires the team's capability \(\alpha_3 \geq \gamma\). When \(\gamma\) is set larger, more agents get involved in the optimization and the problem space is larger. We do not scale noncumulative capabilities. For one task, the requirements on different capability types are imposed with `and' logic.
These tasks are distributed in a city. An example of 16 tasks (two tasks from each type) is shown in Fig. \ref{fig:medical_task_distribution}. The agents start from the base, which, in practice, can be a hospital.

We use the M3500 dataset \cite{olson2006fast} as the city's road map in this illustration.
We assume that we have a viral exposure level (a real number) at some locations and can infer a cost map of the virality level using Gaussian process \cite{williams2006gaussian, fu2020heterogeneous} based on the assumption that the neighborhood of a high-cost location is also high-cost.
For example, in the city map in Fig. \ref{fig:medical_task_distribution}, we randomly choose 6 contaminated and 6 proven-safe regions as samples and learn a viral exposure map. The blue and red regions have low and high cost, respectively, while the white regions have less information, high ambivalence, and are assigned a medium cost. Based on this, we compute a viral-exposure-based travel cost (A* path \cite{pohl1973avoidance}) between the tasks and regard it as the energy cost for the edges in the graph in Fig. \ref{fig:graphical_model}.

We choose the agents and tasks from their types in TABLEs \ref{tab:medical_vehicle_capability} and \ref{tab:medical_task_requirement} and compose 32 mission cases where the agent numbers, task numbers, and the coefficient \(\gamma\) are chosen from \{21, 70, 140\}, \{16, 24, 32, 40\}, and \{1, 3, 5, 10\}, respectively.
For each mission case, we generate 6 instances with randomly selected tasks locations, which result in different graphical models in Fig. \ref{fig:graphical_model}. For the test cases, task \(i\) and \({i+8}\) have the same requirements, but are at different locations.

Based on these cases, we evaluate the mission performance and computational cost of the three models in TABLE \ref{tab:three_models}:
\begin{itemize}
    \item \modelrisk{}: The risk minimization model
    \item \modeldet{}: The risk part, \(h_i\), is removed from the objective function \eqref{eqn:nonlin_penalty_objective}.
    \item \modelint{}: Solve the same problem as \modeldet{} without the flow decomposition. The size of the math problem is larger and harder to solve. This is our baseline from \cite{fu2020heterogeneous} to compare.
\end{itemize}

\newlength{\tasktabwid}
\setlength{\tasktabwid}{0.02\linewidth}

\begin{table}[t]
  \centering
  \caption{Definitions of the capability types.}
  \label{tab:capability_type}%
    \begin{tabular}{ll}
    \toprule
    Cap & Definition \\
    \midrule
    \(1\)     & Fly (noncumulative) \\
    \(2\)     & Equipped with a freezer \\
    \(3\)     & Deliver materials \\
    \(4\)     & Conduct perception \\
    \(5\)     & Remove and collect harmful materials \\
    \(6\)     & Conduct viral test \\
    \(7\)     & Physically interact and conduct treatment \\
    \(8\)     & Spray disinfectant \\
    \(9\)     & Place signals and barricades to conduct quarantine enforcement \\
    \bottomrule
    \end{tabular}%
\end{table}%

\begin{table}[t]
  \centering
  \caption{Stochastic agent capabilities.}
  \label{tab:medical_vehicle_capability}%
    \begin{tabular}{lp{\tasktabwid}p{\tasktabwid}p{\tasktabwid}p{\tasktabwid}p{\tasktabwid}p{\tasktabwid}p{\tasktabwid}p{\tasktabwid}p{1.2\tasktabwid}}
    \toprule
    Agent species & \(1\)     & \(2\)     & \(3\)     & \(4\)     & \(5\)     & \(6\)     & \(7\)     & \(8\)     & \(9\) \\
    \midrule
    \(1\): Quadcopter & 1     &       & 1     & 1     &       &       &       & 1     &   \\
    \(2\): Vehicle &       &       & 1     &       &       &       &       &       & 2 \\
    \(3\): Vehicle (freezer) &       & 1     & 1     &       &       &       &       &       &   \\
    \(4\): Vehicle (contaminants) &       &       &       &       & 1     &       &       & 1     &   \\
    \(5\): Guidance robot &       &       &       & 1     &       &       &       &       & 5 \\
    \(6\): Test robot &       &       &       &       &       & 1     &       &       &   \\
    \(7\): Treatment robot &       &       &       &       &       & 1     & 1     &       &   \\
    \bottomrule
    \end{tabular}%
\end{table}%

\begin{table}[t]
  \centering
  \caption{Stochastic task requirements.}
  \label{tab:medical_task_requirement}%
    \begin{tabular}{l@{\hskip -0.005\linewidth} p{\tasktabwid}p{\tasktabwid}p{\tasktabwid}p{\tasktabwid}p{\tasktabwid}p{\tasktabwid}p{\tasktabwid}p{\tasktabwid}p{1.2\tasktabwid}}
    \toprule
    Task type & \(1\)     & \(2\)     & \(3\)     & \(4\)     & \(5\)     & \(6\)     & \(7\)     & \(8\)     & \(9\) \\
    \midrule
    \(1\): Goods delivery &       &       & \(\gamma\)     &       &       &       &       &       &  \\
    \(2\): Goods delivery (fly) & \(1\)     &       & \(\gamma\)     &       &       &       &       &       &  \\
    \(3\): Medical kits delivery &       & \(\gamma\)     & \(\gamma\)     &       &       &       &       &       &  \\
    \(4\): Contaminants removal &       &       &       &       & \(\gamma\)     &       &       & \(\gamma\)     &  \\
    \(5\): Open area disinfection &       &       &       & \(\gamma\)     & \(\gamma\)     &       &       & \(2\gamma\)     &  \\
    \(6\): Quarantine enforcement &       &       & \(\gamma\)     & \(\gamma\)     & \(\gamma\)     &       &       & \(\gamma\)     & \(10\gamma\) \\
    \(7\): Viral tests &       &       &       &       &       & \(\gamma\)     &       &       &  \\
    \(8\): Remote treatment &       &       &       &       &       & \(\gamma\)     & \(\gamma\)     &       &  \\
    \bottomrule
    \end{tabular}%
\end{table}%

\begin{table}[t]
  \centering
  \caption{Task descriptions.}
  \label{tab:medical_task_description}%
    \begin{tabular}{lp{0.55\linewidth}}
    \toprule
    Task type & Description \\
    \midrule
    \(1\): Goods delivery & Deliver goods. \\
    \(2\): Goods delivery (fly) & Deliver goods using quadcopters. \\
    \(3\): Medical kits delivery & The team should be equipped with freezers. \\
    \(4\): Contaminants removal &  Remove contaminated materials and then disinfect the location. \\
    \(5\): Open area disinfection &  Perceive a contaminated area to conduct contaminants removal and disinfection.\\
    \(6\): Quarantine enforcement & Perceive, carry materials to, disinfect a contaminated area, and then place signals and barricades to enforce a quarantine.\\
    \(7\): Viral tests & Conduct viral tests.\\
    \(8\): Remote treatment & Conduct viral tests and treatment.\\
    \bottomrule
    \end{tabular}%
\end{table}%

\begin{table}[htb!]
  \centering
  \caption{Three teaming models. `Agent var' stands for `agent variables'.}
  \label{tab:three_models}%
    \begin{tabular}{llll}
    \toprule
    Model & Agent var & Objective & Variable number \\
    \midrule
    \modelint{} \cite{fu2020heterogeneous}  & Binary & Energy + Time & Large \\
    \modeldet{}   & Real & Energy + Time & Relatively small \\
    \modelrisk{}  & Real & Energy + Time + Risk & Relatively small \\
    \bottomrule
    \end{tabular}%
\end{table}%

\subsubsection{Computational Cost and Discussion}\label{sec:medical_experiment_computation}

\begin{figure}[hbt!]
	\centering
	\includegraphics[height=0.60\linewidth, trim=120 190 140 215, clip]{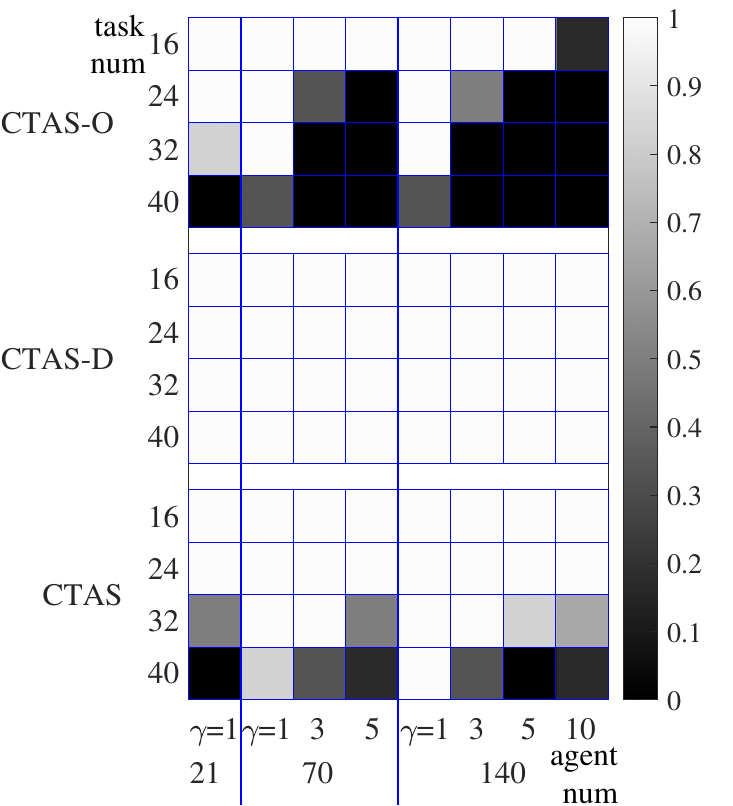}
	\caption{The success rate of the models for each mission test case (6 instances are tested for each case). The grayscale colors correspond to the value specified in the color bars on the right. }
	\label{fig:medical_success_rate}
\end{figure}

\begin{figure}[hbt!]
    \centering
	\subfloat[\label{fig:medical_ori_lin_gap}]{
    	\includegraphics[height=0.60\linewidth, trim=120 190 140 215, clip]{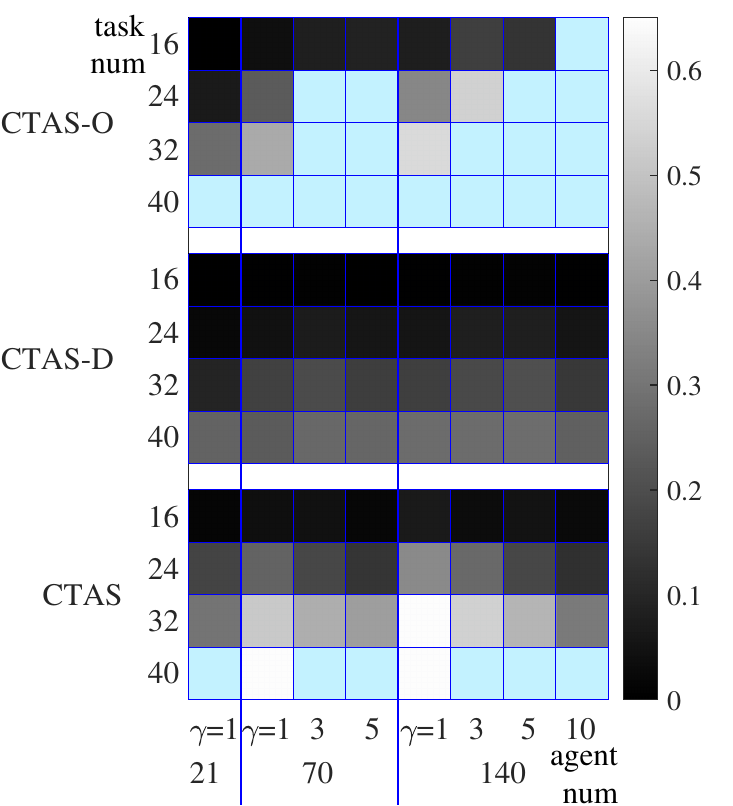}}
	\subfloat[\label{fig:medical_d_lin_gap}]{
    	\includegraphics[height=0.60\linewidth, trim=210 190 140 215, clip]{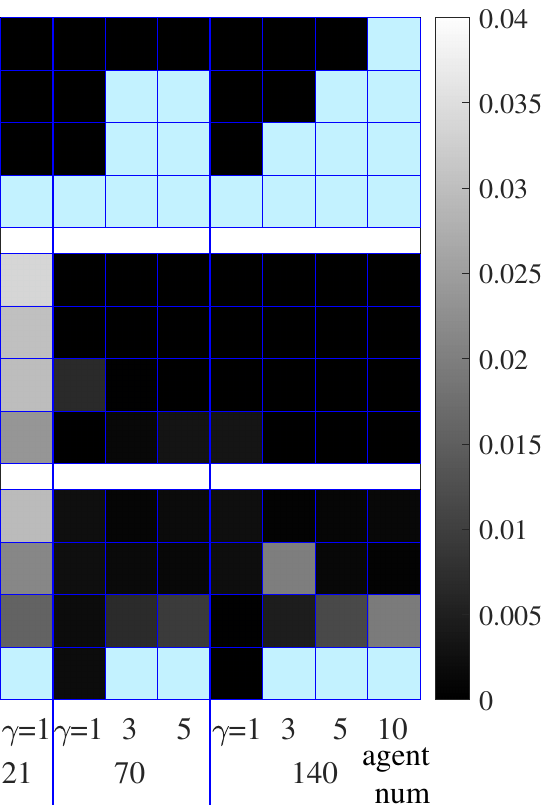}}
    \hfill
	\caption{The optimality gap of the three models applied to the 32 test cases, defined as {(objective value - lower bound) / lower bound}. A smaller optimality gap indicates a better solution. The grayscale colors correspond to the value specified in the color bars on the right. The \textbf{blue cells} mean the solver's success rate for the specific mission case is \(<50\%\) for the time limit picked (120 seconds). (a) The optimality gap before the flow rounding process. The two white cells in the \modelrisk{} group are outliers whose values are around 1.0. (b) The increased optimality gap due to the rounding process.}
	\label{fig:medical_three_models}
\end{figure}

\begin{figure}[hbt!]
	\centering
	\includegraphics[height=0.60\linewidth, trim=120 190 140 215, clip]{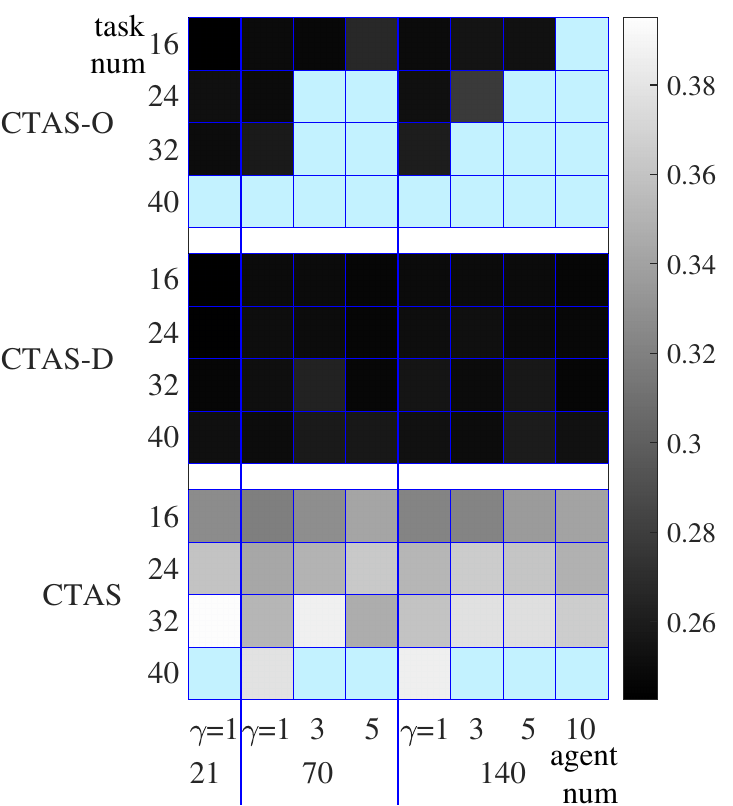}
	\caption{The mean probability of success for the tasks using the three models.}
	\label{fig:medical_mean_prob}
\end{figure}

\begin{figure}[hbt!]
    \centering
    \subfloat[]{
    	\includegraphics[height=0.23\linewidth, trim=115 290 140 325,  clip]{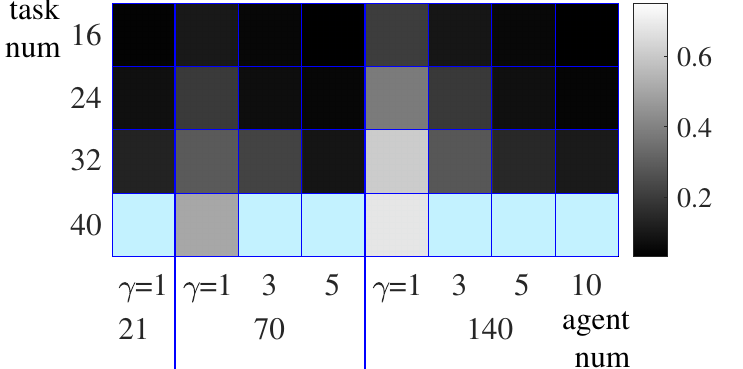}} 
    \subfloat[]{
    	\includegraphics[height=0.23\linewidth, trim=115 290 140 325, clip]{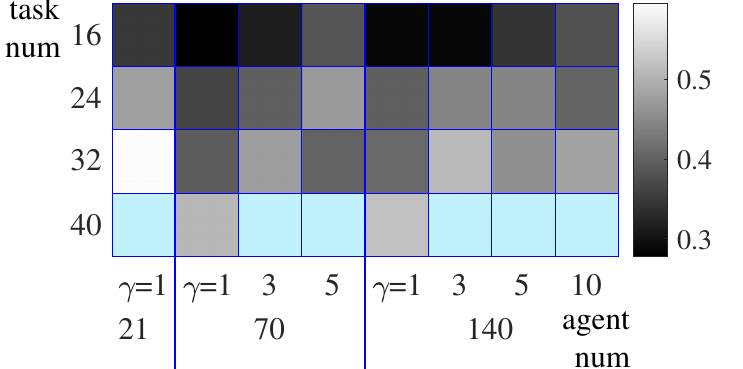}}
	\hfill
	\caption{Comparing the results of \modelrisk{} to \modeldet{}. The values (colors) in the girds are (a) increased energy (relative) and (b) increased mean probability (relative) by adding the risk as an objective. I.e., the values are (\modelrisk{} \(-\) \modeldet{}) / \modeldet{}.}
	\label{fig:medical_risk_vs_det}
\end{figure}

\begin{figure}[hbt!]
	\centering
	\includegraphics[width=0.52\linewidth, trim=120 280 140 320, clip]{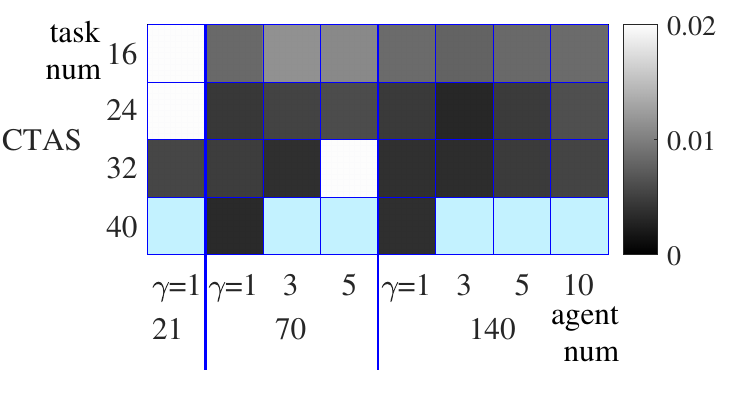} 
	\caption{The relative approximation gap of the nonlinear CVaR through sampling and linear programming.}
	\label{fig:medical_risk_approx_gap}
\end{figure}

A 120-second time limitation is added to the solvers of the models for all test cases.
For each mission case, we run the 6 test instances and if the solver can find a feasible solution within the time limit, we consider it successful. The success rate of each model is shown in Fig. \ref{fig:medical_success_rate}.
We show the detailed average performances for the mission sizes with success rates larger than 50\% in Fig. \ref{fig:medical_three_models}-\ref{fig:medical_risk_approx_gap}.
The optimality gaps after the rounding process in Sec. \ref{sec:flow_round_up} are given in Fig. \ref{fig:medical_ori_lin_gap}. 
The blue cells mean the solver's success rate for the specific mission size is less than 50\% for the time limit picked.
According to the two figures, \modeldet{} and \modelrisk{} can solve much larger teaming problems than \modelint{}. The largest problem that \modeldet{} solves in this example involves 140 agents and 40 tasks.

The increases in the optimality gaps due to the rounding process are shown in Fig. \ref{fig:medical_d_lin_gap}. For most of the cases, the rounding process introduces an increase in the optimality gap within 1\%. The leftmost cases have larger increases because the agent numbers are small, and rounding has a larger overall influence. Since \modelint{} involves no rounding process, the increased gap is 0 for all test cases.

In these test cases, the agent capabilities are Gaussian distributed. Therefore, given the estimation of the means and variances, we can calculate the probability of success and its mean as follows
\begin{align}
    P (\text{task } i \text{ succeeds}) =& \underset{\text{capability } a}{\prod} P(a \text{ satisfied}) \quad \forall i \in M \nonumber \\ 
    \text{mean } P(\text{success}) =& \left(\underset{i \in M}{\prod} P (\text{task } i \text{ succeeds}) \right)^{1/n_m}  \nonumber
\end{align}

The mean probabilities of success for the three models are shown in Fig. \ref{fig:medical_mean_prob}. For all cases, \modelint{} and \modeldet{} models roughly result in a probability of 0.25. This is because each task requires two capabilities on average. When the risk is not considered, these two models tend only to match the expected requirement, and the probability of matching a single capability is 0.5. 
Clearly, the risk minimization model increases the probability of success for the tasks.

In Fig. \ref{fig:medical_risk_vs_det}, we compare the result of \modelrisk{} to its deterministic version \modeldet{}. With the chosen \(\beta\)\footnote{\(\beta\): hyperparameter in the CVaR function \(\eta_\beta(\cdot)\) in Fig. \ref{fig:cvar_definition} and objective function \eqref{eqn:nonlin_cvar_objective_h_min} and \eqref{eqn:nonlin_cvar_objective_h_sum}.},
\(C_e\), and \(C_h\)\footnote{\(C_e\) and \(C_h\): the penalty coefficients on energy cost and task completion in equation \eqref{eqn:nonlin_penalty_objective}.}, for most of the cases, a \(\sim\)20\% increase in energy cost introduces a  \(\sim\)35\% increase in the mean probability of success.

\begin{figure}[hbt!]
	\centering
	\includegraphics[width=0.8\linewidth, trim=75 240 100 250, clip]{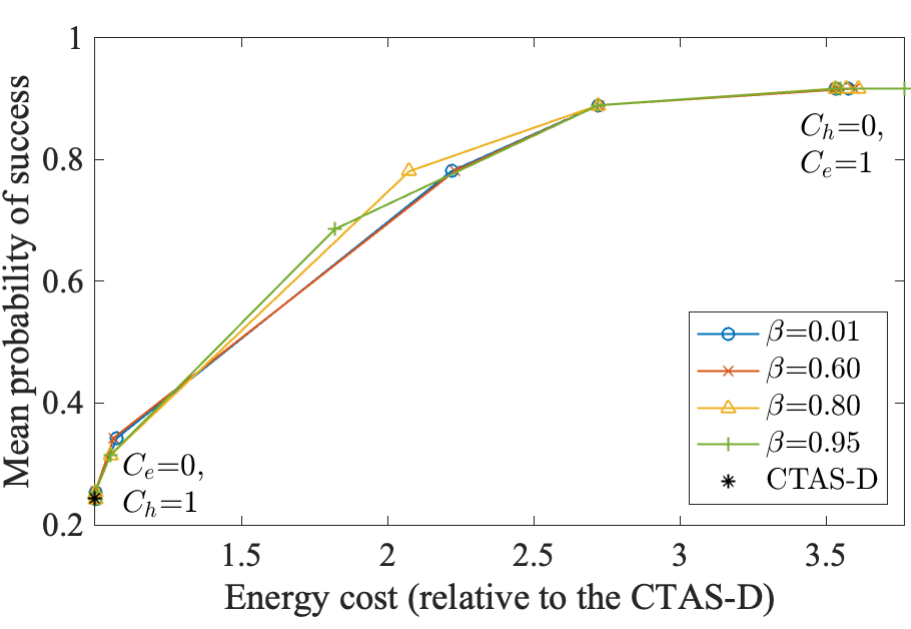}
	\caption{Trade-off between energy cost and risk of task's non-completion for a case with \{16 tasks, 21 agents, and \(\gamma = 1\)\}. The upper-left corner is the direction of the Pareto set.}
	\label{fig:medical_prob_eng_tradeoff}
\end{figure}

The trade-off between energy cost and robustness in the objective function and can be tuned smoothly.
The trade-off gained by changing the \(C_e\), \(C_h\), and \(\beta\) is shown in Fig. \ref{fig:medical_prob_eng_tradeoff}, using the smallest test case as an example. The penalty coefficient has a major impact on the trade-off between energy and the probability of success, while \(\beta\) has a local impact on the trade-off. According to our investigation, choosing \(\beta = 0.8 - 0.97\) would mostly cover the Pareto set. This shows the advantage of using the CVaR as a risk metric: better Pareto optima are gained than optimizing the expectation (\(\beta \rightarrow 0\)).

\newcommand{\medicalfolder}{figure/medical_revise} 

For the sample average approximation of the CVaR, we use 500 samples to approximate the Gaussian distributions. Given the solution, we compare the nonlinear objectives to their sample approximations and find that the relative approximation gaps are smaller than \(1\%\) for most of the cases, as shown in Fig. \ref{fig:medical_risk_approx_gap}. This shows that the approximation quality is good using 500 samples.

\subsubsection{Mission Performance and Discussion}\label{sec:medical_experiment_performance}

This section uses a test case with 16 tasks, 21 agents, and \(\gamma = 1\) as an example to compare the solutions generated by the three models. 
In TABLE \ref{tab:medical_picked_result}, we list the performance metrics of 5 tasks whose success rate is increased by the \modelrisk{} model. All three models are solved to optimal within the time limit. The models presented in this paper end up with much fewer variables for the same problem. By comparing \modelrisk{} with \modeldet{}, we see that the risk minimization model reduces the CVaR and increases the probability of success for 5 tasks out of 16, with a small increase (5.3\%) in the overall energy cost.

The teams for the last 8 tasks are shown in TABLE \ref{tab:medical_picked_teams}.
In the table, \(v_i\) stands for one agent from species \(i\).
As expected, the \modelrisk{} model puts more agents in the team to reduce the CVaR. This task assignment results from simultaneously considering the energy cost. As an example, \modelrisk{} puts more agents in the team of tasks \(9\)-\({12}\), but not tasks \(1\)-\(4\), even if they are of the same task types. Because energy and time costs are jointly considered, and tasks \(1\)-\(4\) are far from the depot, adding more agents to ensure redundancy and robustness could potentially result in much higher costs. The risk minimization model also generates the routes and a consistent schedule. For example, one agent from species \(1\), \(4\), and \(5\) visit task \({13}\) at the same time as a team. The routes of species \(1\), \(4\), and \(5\) are shown in Fig. \ref{fig:medical_traj}. These routes minimize overall travel distances and avoid the high-cost red regions to lower energy costs.

In summary, according to the teams in TABLE \ref{tab:medical_picked_teams} and routes in Fig. \ref{fig:medical_traj}, the \modelrisk{} framework generates a consistent schedule for coordination, outputs routes that minimize energy costs, and assigns tasks to agents such that redundancy is preserved at low costs to ensure a higher probability of task completion. The computational evaluation in Sec. \ref{sec:medical_experiment_computation} (particularly, Fig. \ref{fig:medical_three_models}) shows that the frameworks \modelrisk{} and \modeldet{} scale to 140 agents and 40 tasks with low optimality gaps. The scalability in agent number is better than the task number. Furthermore, \modeldet{} still shows no optimality gap degeneration dealing with the largest test case we tested.

\begin{table}[t]
  \centering
  \caption{A comparison of the results of the three models.
  }
  \label{tab:medical_picked_result}%
    \begin{tabular}{llll}
    \toprule
    Item  & \modelint{}   & \modeldet{}   & \modelrisk{} \\
    \midrule
    Variables & 6194  & 4276  & 4384 \\
    Task CVaR & 5.41 \(\times 10^3\) & 5.41 \(\times 10^3\) & -1.03 \(\times 10^4\) \\
    Energy cost & 2.06 \(\times 10^5\) & 2.06 \(\times 10^5\) & 2.17 \(\times 10^5\) \\
    \(P(\text{task }  {9}) \)    & 0.5 & 0.5 & 1 \\
    \(P(\text{task }  {10})\)    & 0.25 & 0.25 & 0.5  \\
    \(P(\text{task }  {11})\)    & 0.25 & 0.25 & 1 \\
    \(P(\text{task }  {12})\)    & 0.25 & 0.25 & 0.5  \\
    \(P(\text{task }  {13})\)    & 0.125 & 0.125 & 0.25 \\
    \bottomrule
    \end{tabular}%
\end{table}%

\begin{table}[t]
  \centering
  \caption{Team configurations given by the three models.
  }
  \label{tab:medical_picked_teams}%
    \begin{tabular}{llll}
    \toprule
    Task  & \modelint{}   & \modeldet{}   & \modelrisk{} \\
    \midrule
    \(9\)  & \(v_3    \) & \(v_3    \) & \(v_1 \times 3, v_2 \times 3, v_3 \times 3 \) \\
    \({10}\)  & \(v_1    \) & \(v_1    \) & \(v_1 \times 2 \) \\
    \({11}\)  & \(v_3    \) & \(v_3    \) & \(v_1 \times 2, v_2 \times 3, v_3 \times 3 \) \\
    \({12}\)  & \(v_4    \) & \(v_4    \) & \(v_1 \times 2, v_4 \) \\
    \({13}\)  & \(v_1, v_4    \) & \(v_1, v_4    \) & \(v_1, v_4, v_5 \) \\
    \({14}\)  & \(v_2 \times 3, v_4, v_5\) & \(v_2 \times 3, v_4, v_5\) & \(v_2 \times 3, v_4, v_5\) \\
    \({15}\)  & \(v_7\) & \(v_7\) & \(v_7\) \\
    \({16}\)  & \(v_7\) & \(v_7\) & \(v_7\) \\
    \bottomrule
    \end{tabular}%
\end{table}%

%% file: section/conclusion.tex
\section{Conclusions and Future Work}\label{sec:conclusion}

This paper addresses a complex task allocation problem in the category of CD-[ST-MR-TA].
We propose a mixed-integer programming model that simultaneously optimizes the task decomposition, assignment, and scheduling. The uncertainty within the team's capability is considered through risk minimization, and a robust metric, conditional value at risk (CVaR), is minimized to ensure the robustness to task completion (enough capability are assigned). The framework contributes to a domain-independent representation for complex tasks and heterogeneous agent capabilities that can generalize to multi-agent applications where the major goals are satisfying task-required capabilities. A two-step solution method is described, and the whole framework is evaluated in two different practical test cases. Results show that the framework scales up to 140 agents and 40 tasks for the cases tested and solves the problems with low optimality gaps. Given the selected hyper-parameters, the resulting assignments and schedules provide a reasonable trade-off between energy, time, and the probability of success. The task assignment performance (apart from the scheduling) is also demonstrated through a comparison with the STRATA framework in the capture the flag case.

Future work will consider an extension of this work to a distributed framework, which could further improve the scalability of the system and the robustness to communication constraints and the loss of agents.
In addition, a probabilistic learning method that automatically estimates the parameters in the representation of task requirements and agent capabilities from current and previous task executions is an interesting future work. Such a learning method would enable the possibility of closing the loop of the task assignment and scheduling, and iteratively improving the performance.
For modeling choices, we will consider imposing a necessary and sufficient energy constraint in the \modelrisk{} model in Sec. \ref{sec:problem_model} while still preserving the scalability of the current framework.

%% file: section/appendix.tex
\appendices
\section{Non-trivialness of the Flow Decomposition}\label{sec:appendix_flow}

A rounding or splitting without solving an optimization can break flow constraints or result in suboptimal solutions.

\begin{figure}[h!]
	\centering
	\subfloat[\label{fig:round_case1}]{
    	\includegraphics[width=0.8\linewidth, trim=0 0 0 0, clip]{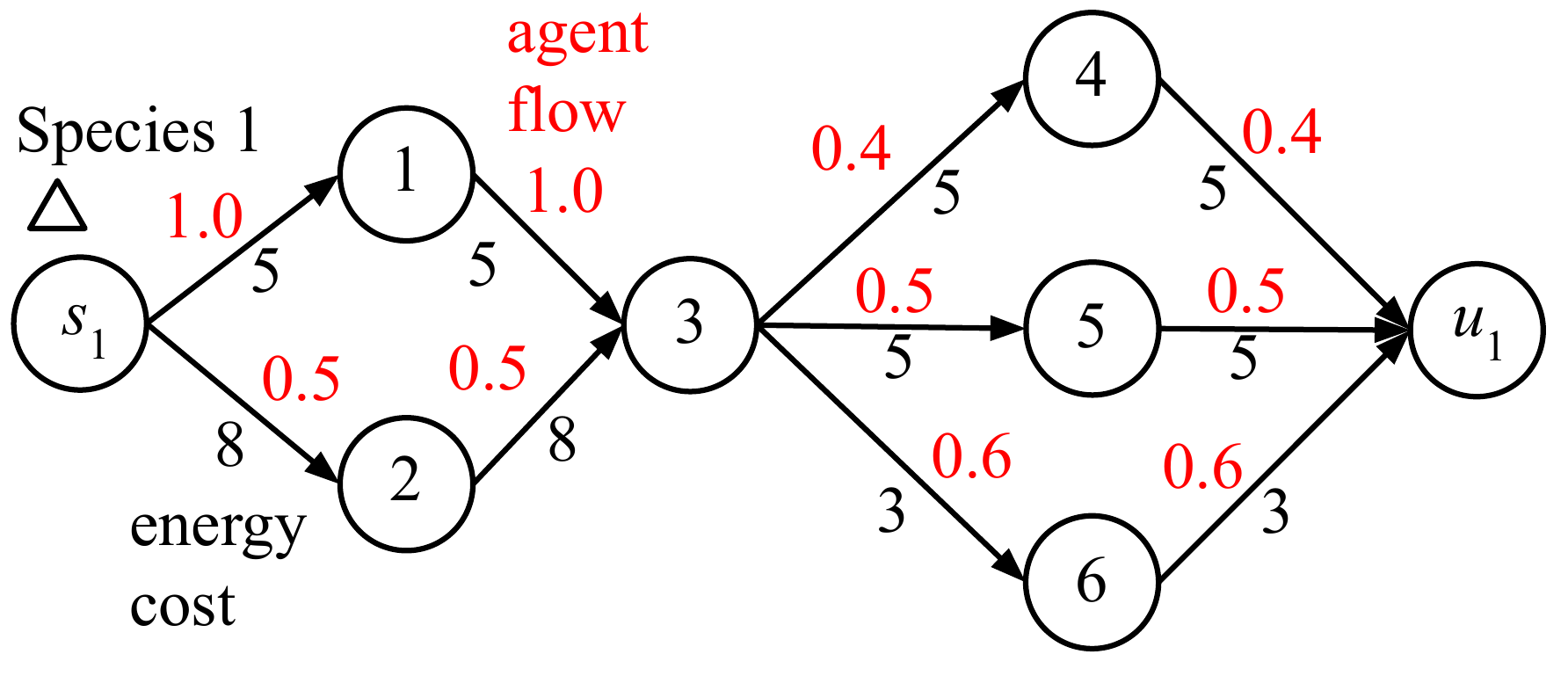}}
    \hfill
	\subfloat[\label{fig:round_case2}]{
    	\includegraphics[width=0.8\linewidth, trim=0 0 0 0, clip]{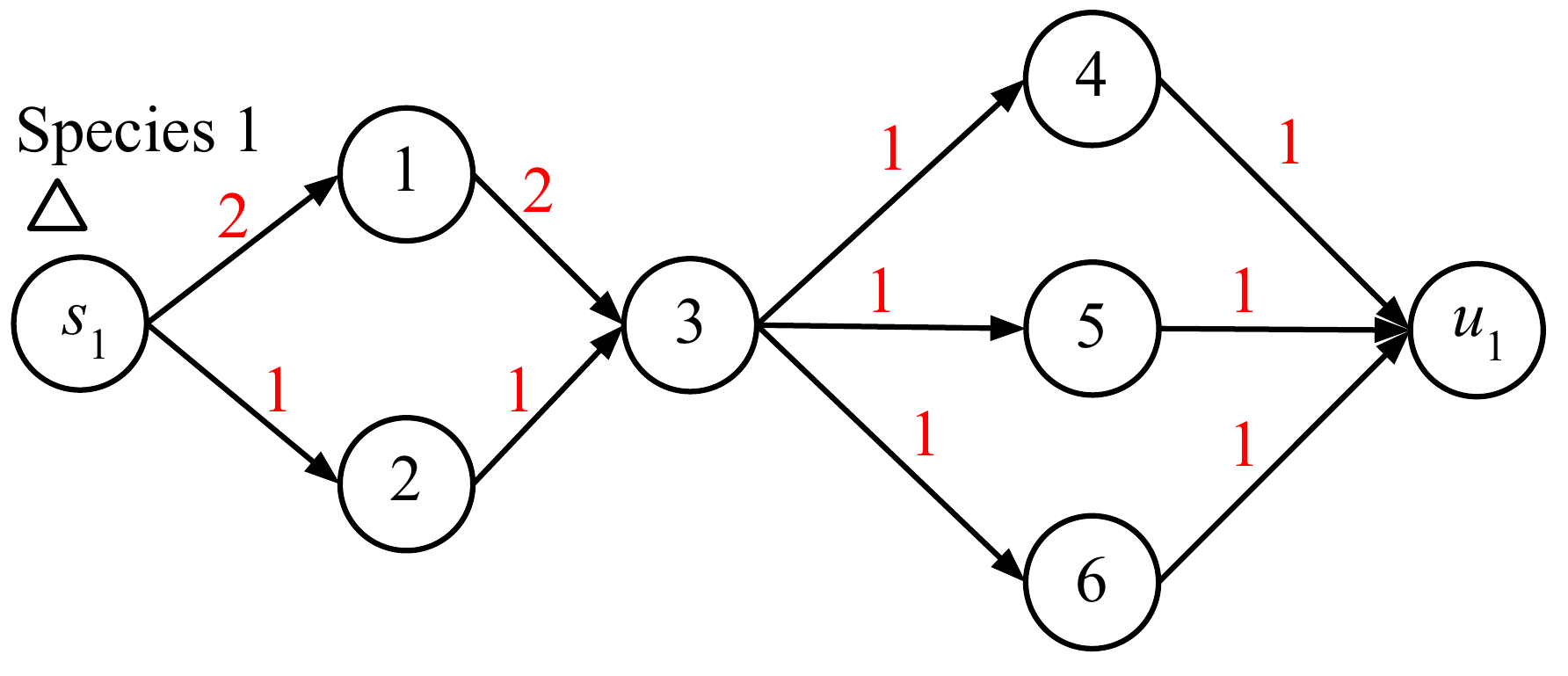}}
    \hfill
	\subfloat[\label{fig:round_case3}]{
    	\includegraphics[width=0.8\linewidth, trim=0 0 0 0, clip]{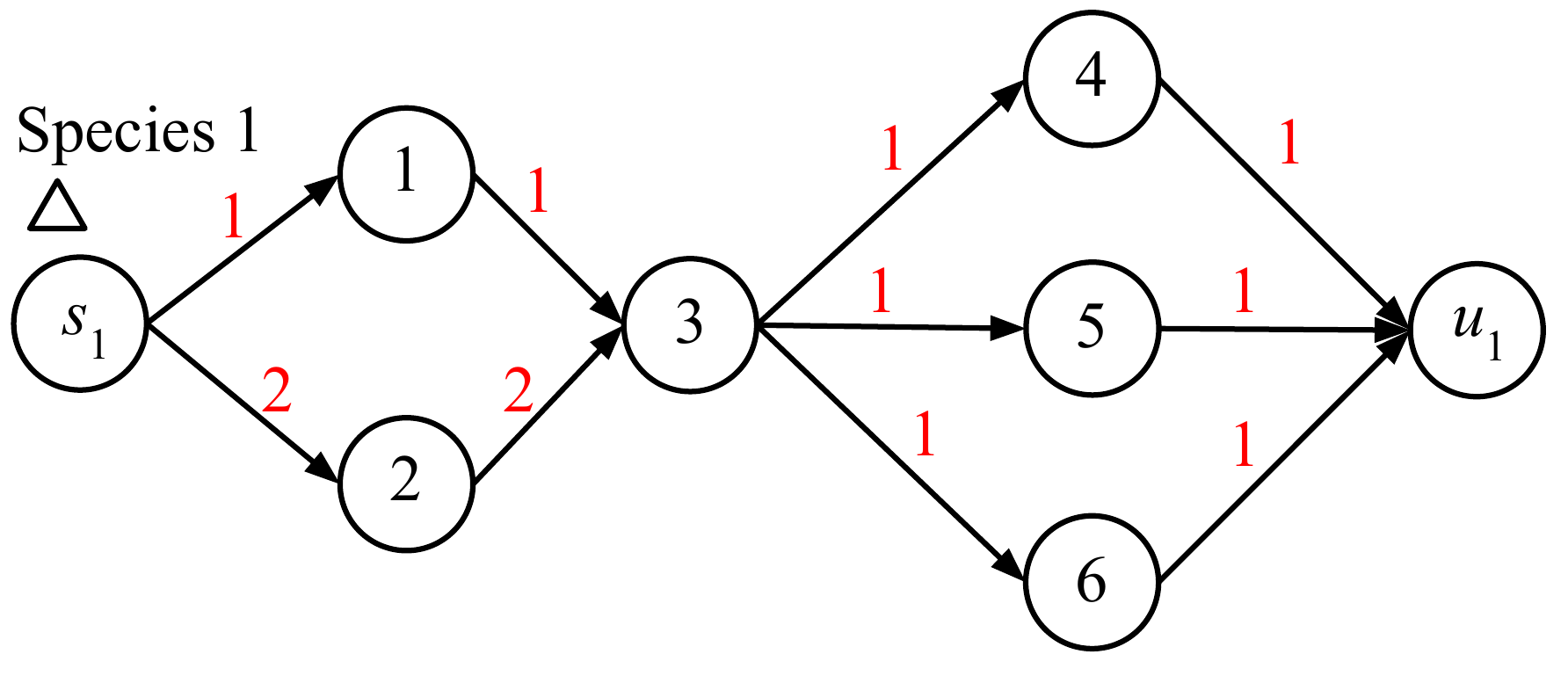}}
    \hfill
	\caption{Round the flow of agent species \(1\). The red number is the agent flow on an edge. The black number below an edge is the energy cost of the edge. (a) The original flow with real numbers. (b) Rounding choice 1. (c) Rounding choice 2.}
	\label{fig:round_case}
\end{figure}

The rounding process has to maintain the flow constraints: for all nodes, the incoming flow must equal the outgoing flow. A naive rounding might break the flow constraint. Take the node \(3\) in Fig. \ref{fig:round_case1} as an example: naively rounding up the flows on an edge will result in 2 incoming agents, but 3 outgoing agents.

The rounding selection process is not unique, even when meeting the flow constraints.
For instance, Fig. \ref{fig:round_case}b-c are both valid ways to round the flow in Fig. \ref{fig:round_case1}. However, the energy cost of the flow in Fig. \ref{fig:round_case3} is higher. Therefore, the optimal rounding process will need to maintain the flow constraint and introduce the lowest additional energy cost.

\begin{figure}[h!]
	\centering
	\subfloat[\label{fig:cover_case1}]{
    	\includegraphics[width=0.8\linewidth, trim=0 0 0 0, clip]{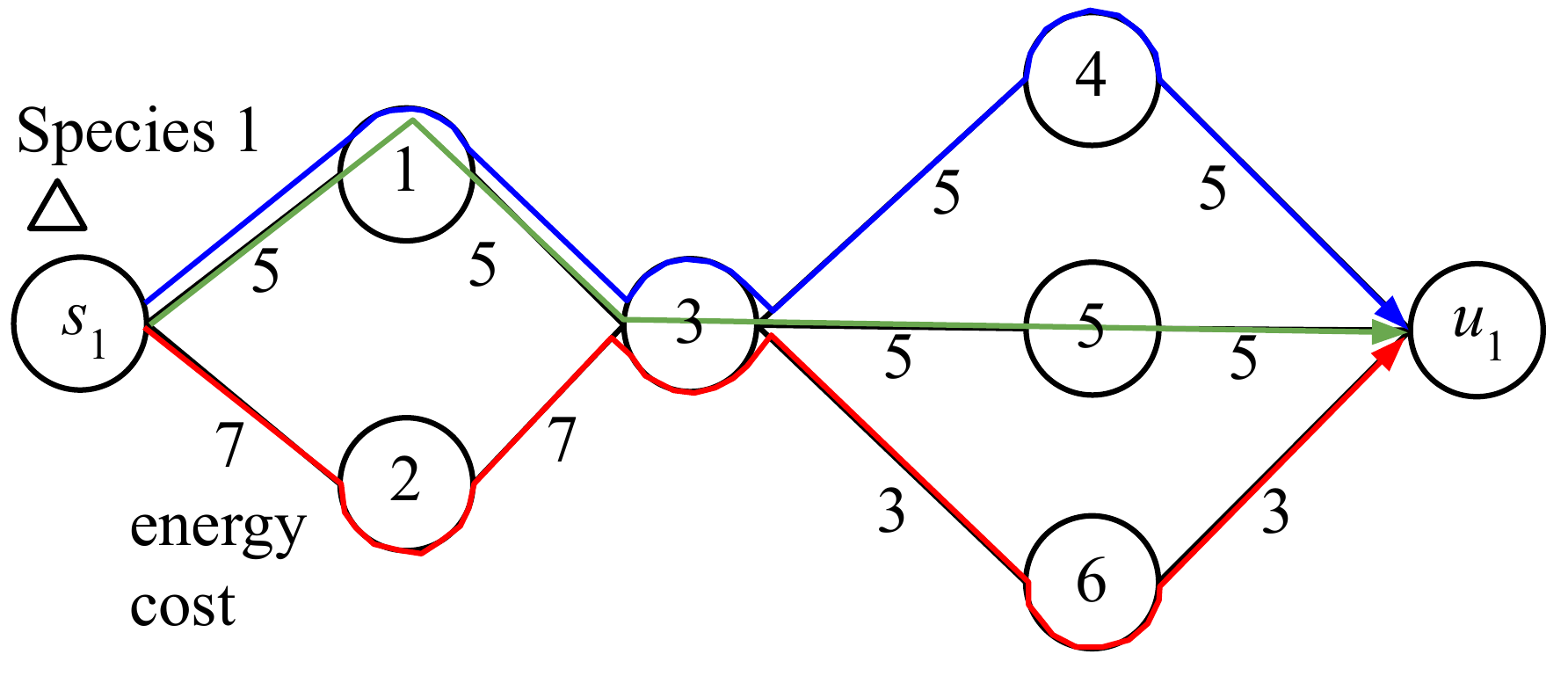}}
    \hfill
	\subfloat[\label{fig:cover_case2}]{
    	\includegraphics[width=0.8\linewidth, trim=0 0 0 0, clip]{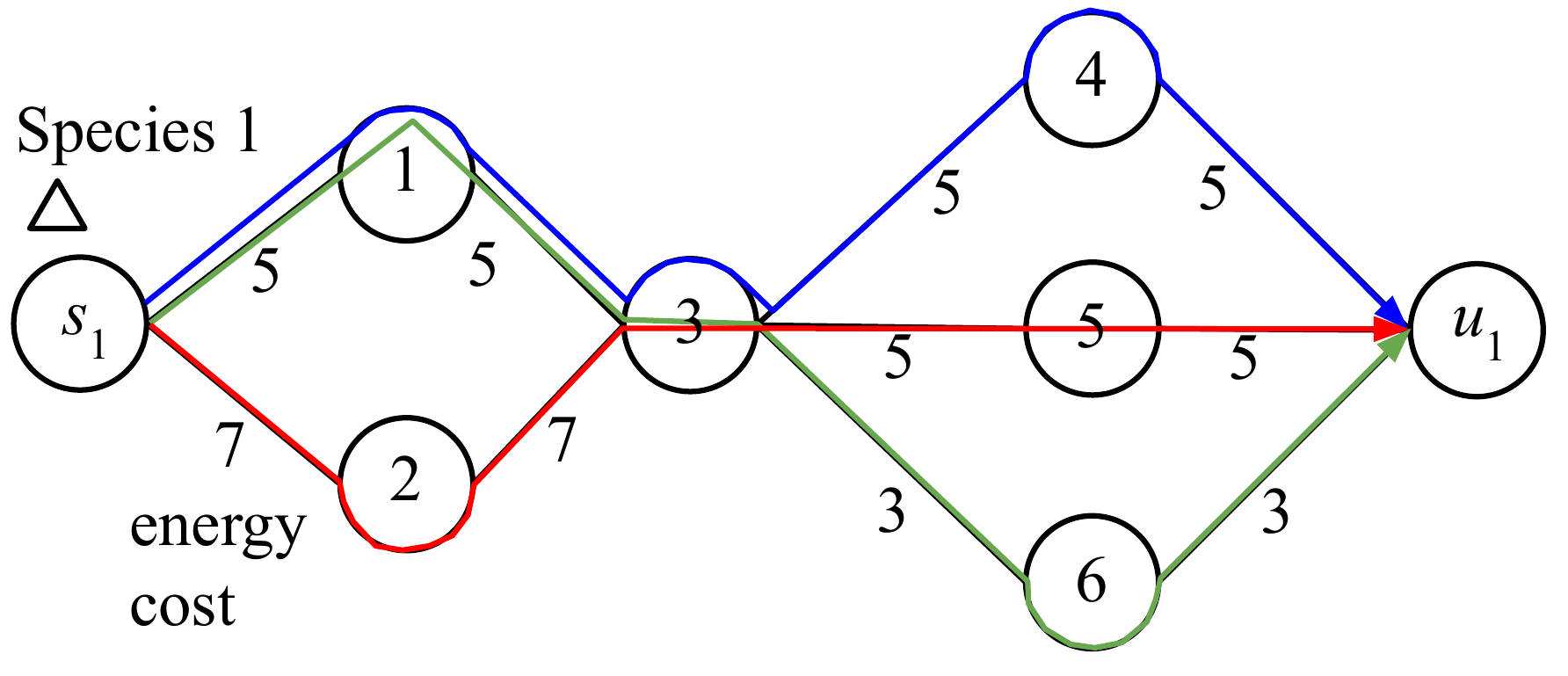}}
    \hfill
	\caption{Split the agent flow in Fig. \ref{fig:round_case2} into 3 paths (colored). The black number below an edge are the energy cost of the edge. There are two choices with different individual costs. (a) Choice 1. (b) Choice 2.}
	\label{fig:cover_case}
\end{figure}

After rounding the flow to integers, there are multiple choices to assign individual agent paths. We call this `splitting the flow into paths'. For example, the flow in Fig. \ref{fig:round_case2} can be split into the 3 agent routes either in Fig. \ref{fig:cover_case1} or Fig. \ref{fig:cover_case2}. Though the total energy costs are equivalent, the energy costs of the three paths in the two choices are \{20, 20, 20\} and \{20, 24, 16\}, respectively. The choice affects the behavior of an individual agent. Here, we prefer the former split, because the energy costs of individual agents are more evenly distributed and the maximum energy cost of an individual agent is smaller. We define the optimal flow split as the set of paths that minimizes the maximum individual energy cost.

\section{Proof of Claim}\label{sec:appendix_integer}

\textbf{Definition 1}: The incidence matrix for a directed graph is a matrix of \(n_\text{node}\) rows and \(n_\text{edge}\) columns: for each row, if an edge enters the node, the matrix value is 1; if an edge leaves the node, the value is -1; otherwise 0.

\textbf{Theorem 1}: The incidence matrix of any directed graph is a totally unimodular (TU) matrix \cite{schrijver2003combinatorial, heller1956extension}.

\textbf{Theorem 2}: If \(A_1\) is a TU matrix and \(E\) is an identity matrix, then \(A=[A_1 | E]\) will be a TU matrix \cite{schrijver2003combinatorial}.

\textbf{Theorem 3}: \(A\) is a TU matrix iff \(A^\transpose{}\) is a TU matrix (by definition of a TU matrix \cite{schrijver2003combinatorial}).

\textbf{Theorem 4}: If \(A\) is a TU matrix and \(\mathbf{b}\) is integer-valued, then each vertex of the polytope defined by \(A \mathbf{x} \leq \mathbf{b}\) is integer-valued \cite{schrijver2003combinatorial}.

\begin{figure}[t]
	\centering
	\includegraphics[width=1\linewidth]{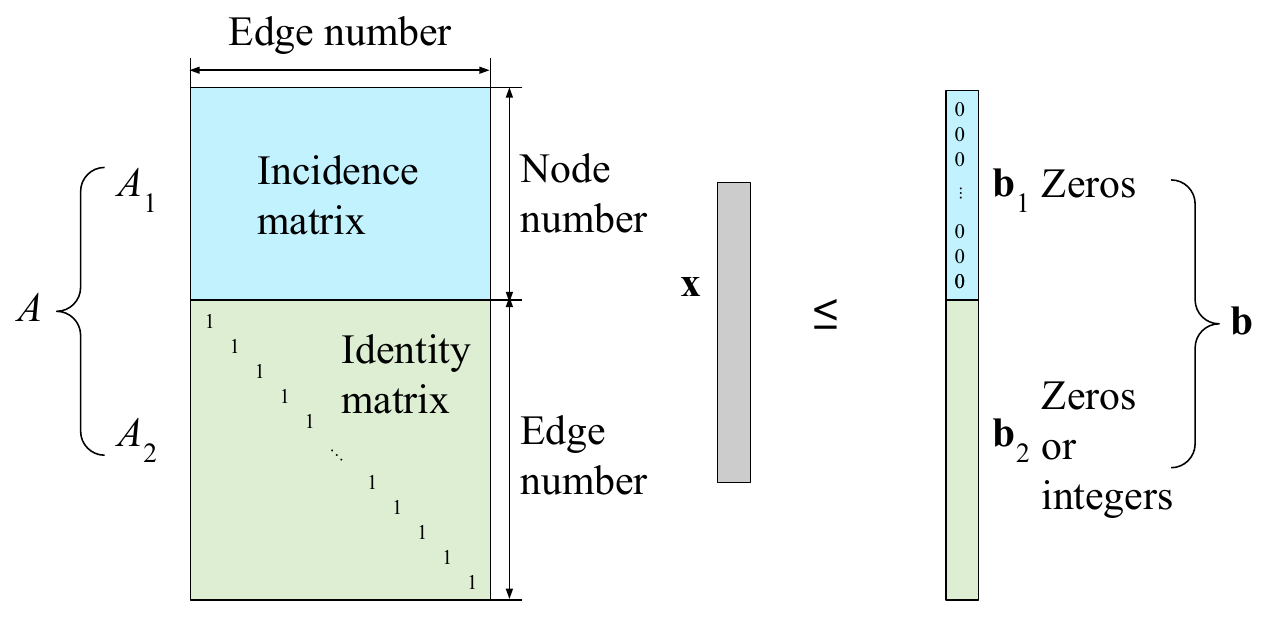}
	\caption{Totally unimodularity of the rounding problem.}
	\label{fig:totally_unimodular_proof}
\end{figure}

\begin{proof}
\textbf{The solutions of the rounding problem from Simplex-based algorithms will be integer-valued.}

Now we write the constraints of the rounding problem, i.e., \eqref{eqn:round_bound}-\eqref{eqn:round_flow_constraint}, in the form of \(A \mathbf{x} \leq \mathbf{b}\). See Fig. \ref{fig:totally_unimodular_proof}, suppose the flow constraint \eqref{eqn:round_flow_constraint} is converted into \(A_1 \mathbf{x} \leq \mathbf{b}_1\), and the bound constraint \eqref{eqn:round_bound} is converted into \(A_2 \mathbf{x}_1 \leq \mathbf{b}_2\).

According to \textbf{Definition 1}, \(A_1\) is an incidence matrix of a directed graph. \(\mathbf{b}_1\) is a zero vector. According to the format of constraints \eqref{eqn:round_bound}, \(A_2\) is an identity matrix, and \(\mathbf{b}_2\) is an integer-valued vector. Therefore, \(\mathbf{b}\) is integer-valued.

According to \textbf{Theorem 1}, \(A_1\) is a TU matrix.

According to \textbf{Theorem 3}, \(A_1^\transpose{}\) is a TU matrix.

Because \(A_2^\transpose{}\) is an identity matrix, according to \textbf{Theorem 2}, \([A_1^\transpose{} | A_2^\transpose{}]\) is a TU matrix.

According to Fig. \ref{fig:totally_unimodular_proof} and \textbf{Theorem 3}, \(A = [A_1^\transpose{} | A_2^\transpose{}]^\transpose{}\) is a TU matrix.

Because matrix \(A\) is a TU matrix and \(\mathbf{b}\) is integer-valued, according to \textbf{Theorem 4}, all vertices of the polytope defined by \(A \mathbf{x} \leq \mathbf{b}\) are integer-valued.

If Simplex-based algorithms are applied to the optimization problem, the algorithms search through vertices of the polytope to find the optimal point.
Therefore, the optimal point must be a vertex of the polytope. Since we have proved that all vertices are integral, the solution to \(\mathbf{x}\) will be integral.
\end{proof}

%% file: section/biography.tex
\begin{IEEEbiography}[{\includegraphics
[width=1in,height=1.25in, trim=0 30 0 0,clip,
keepaspectratio]{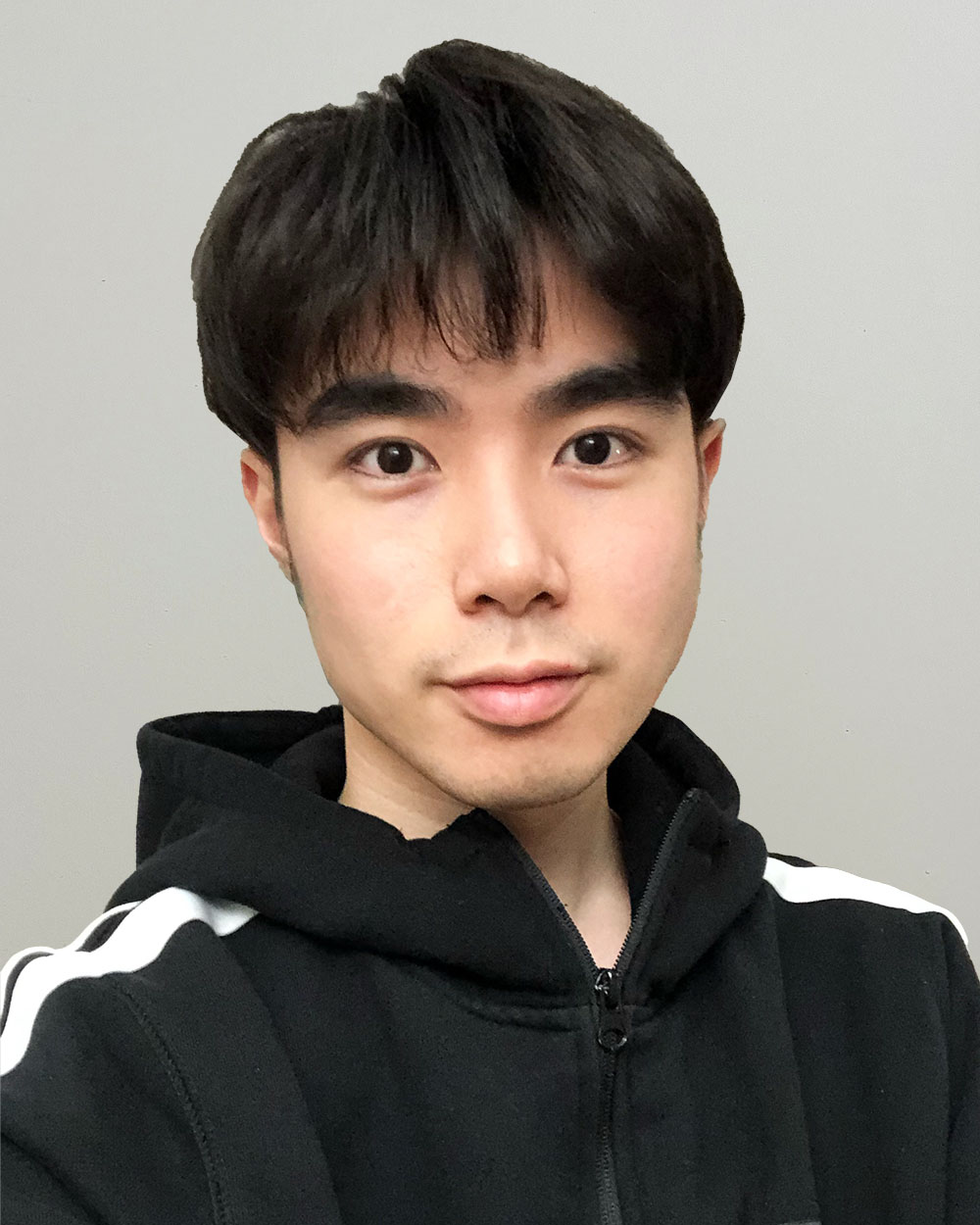}}] 
{Bo Fu}
is currently pursuing a Ph.D. degree at the Department of Robotics, University of Michigan. His current research interest is in the robust and resilient decision-making for heterogeneous robot teams under uncertainty using optimization and learning-based approaches.
He received his M.S. degree from Carnegie Mellon University in 2019, where he worked with visual-inertial odometry systems for quadcopters.
\end{IEEEbiography}

\begin{IEEEbiography}[{\includegraphics
[width=1in,height=1.25in, trim=0 0 0 0,clip,
keepaspectratio]{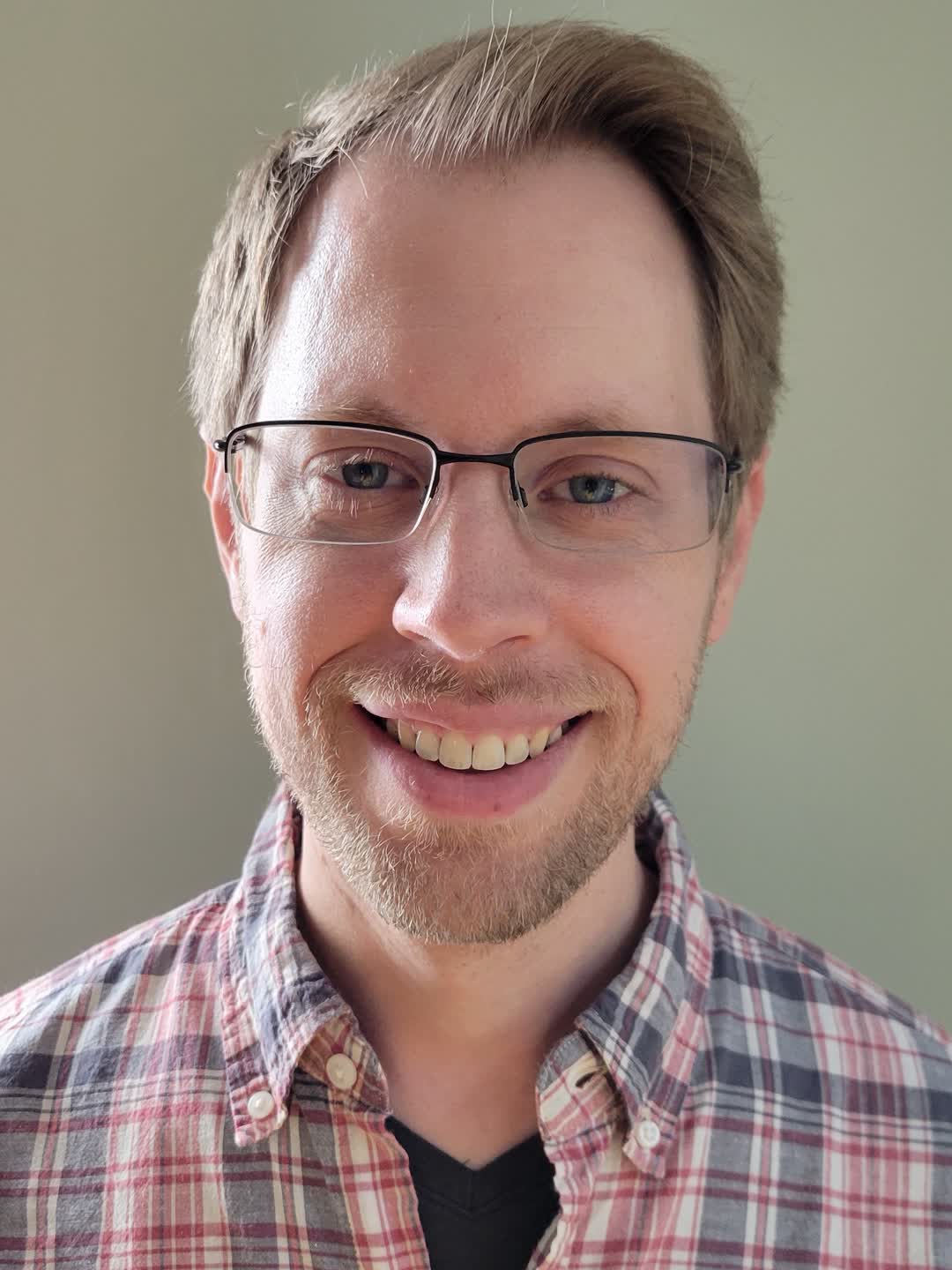}}] 
{William Smith}
received his Ph.D. in mechanical engineering from the University of Michigan in 2014.  He is currently the Autonomous Mobility Technical Specialist within Ground Vehicle Robotics at the U.S. Army's Ground Vehicle Systems Center (GVSC), with he focuses on improving autonomous ground vehicle software through better infrastructure tools, such as simulation, data management, and DevOps.  His research interests include efficient high-fidelity ground-vehicle interaction and vehicle team coordination.
\end{IEEEbiography}

\begin{IEEEbiography}[{\includegraphics
[width=1in,height=1.25in, trim=0 20 0 20,clip,
keepaspectratio]{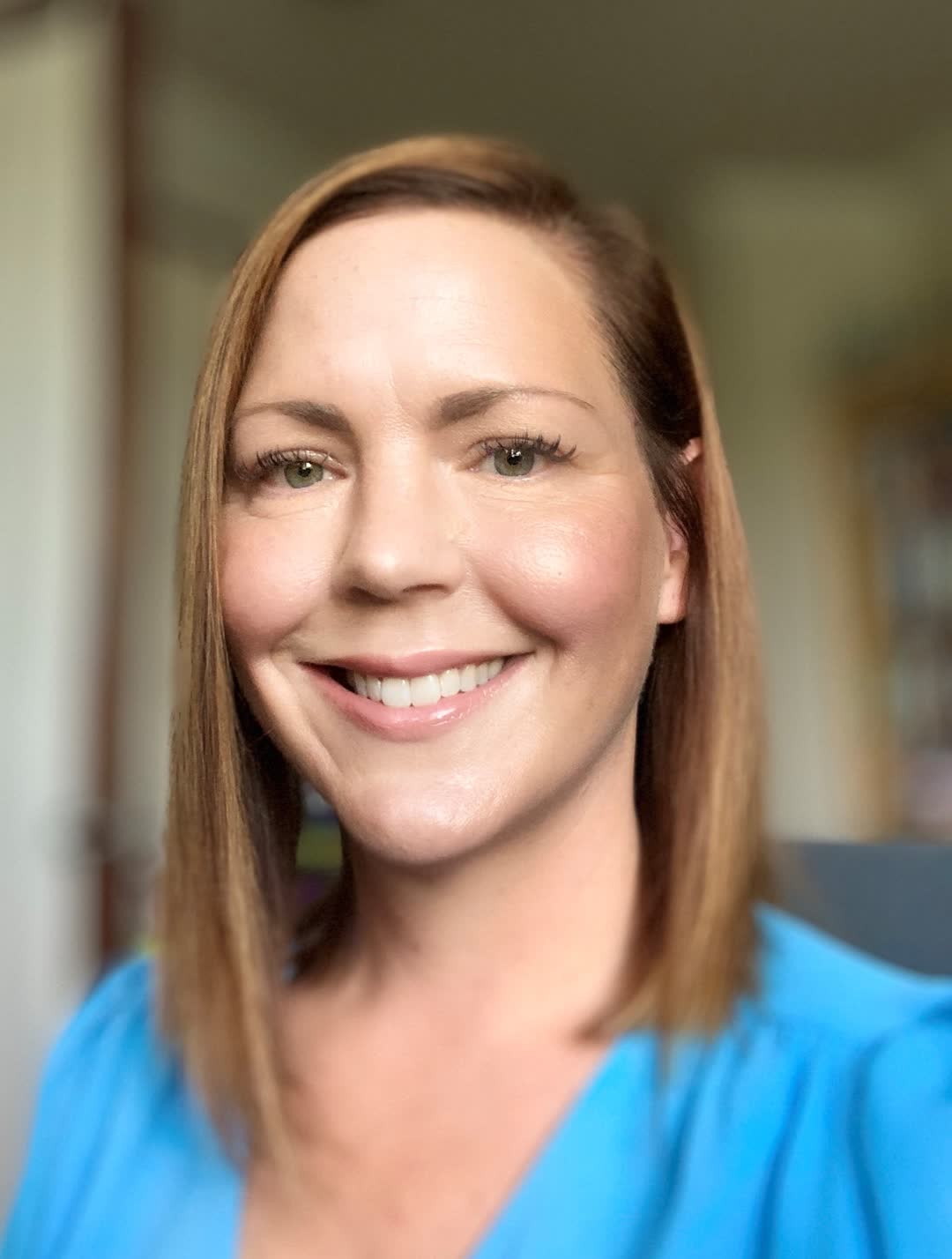}}] 
{Dr. Denise M. Rizzo}
is the Deputy Chief Scientist in the Office of the Chief Scientist at the US Army Ground Vehicle System Center (GVSC). She specializes in modeling, simulation and control of energy intelligent autonomous ground vehicles. Dr. Rizzo received her Ph.D. from Michigan Technological University in 2014. From 2000 through 2008 she was a controls research and development engineer in the Powertrain Group at Chrysler LLC. She joined GVSC in November of 2008 and was promoted to her current position in 2020. Dr. Rizzo has published 40 articles in archival journals, 49 papers in refereed conference proceedings, 4 technical government reports, and holds 2 patents.
\end{IEEEbiography}

\begin{IEEEbiography}[{\includegraphics
[width=1in,height=1.25in,clip,
keepaspectratio]{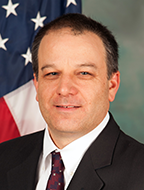}}]
{Dr. Matthew Castanier}
is a Research Mechanical Engineer at the U.S. Army DEVCOM Ground Vehicle Systems Center (GVSC) in Warren, MI. His research interests include powertrain modeling and simulation, multidisciplinary design optimization, and trade space exploration. He received his Ph.D. in Mechanical Engineering from the University of Michigan (UM) in 1995. From 1996 through 2008 he was a member of the research faculty at UM. He joined GVSC in 2008 and was promoted to his current position in 2011. Dr. Castanier has published more than 50 journal articles and over 100 conference papers.
\end{IEEEbiography}

\begin{IEEEbiography}[{\includegraphics
[width=1in,height=1.25in, trim=180 200 200 150,clip,
keepaspectratio]{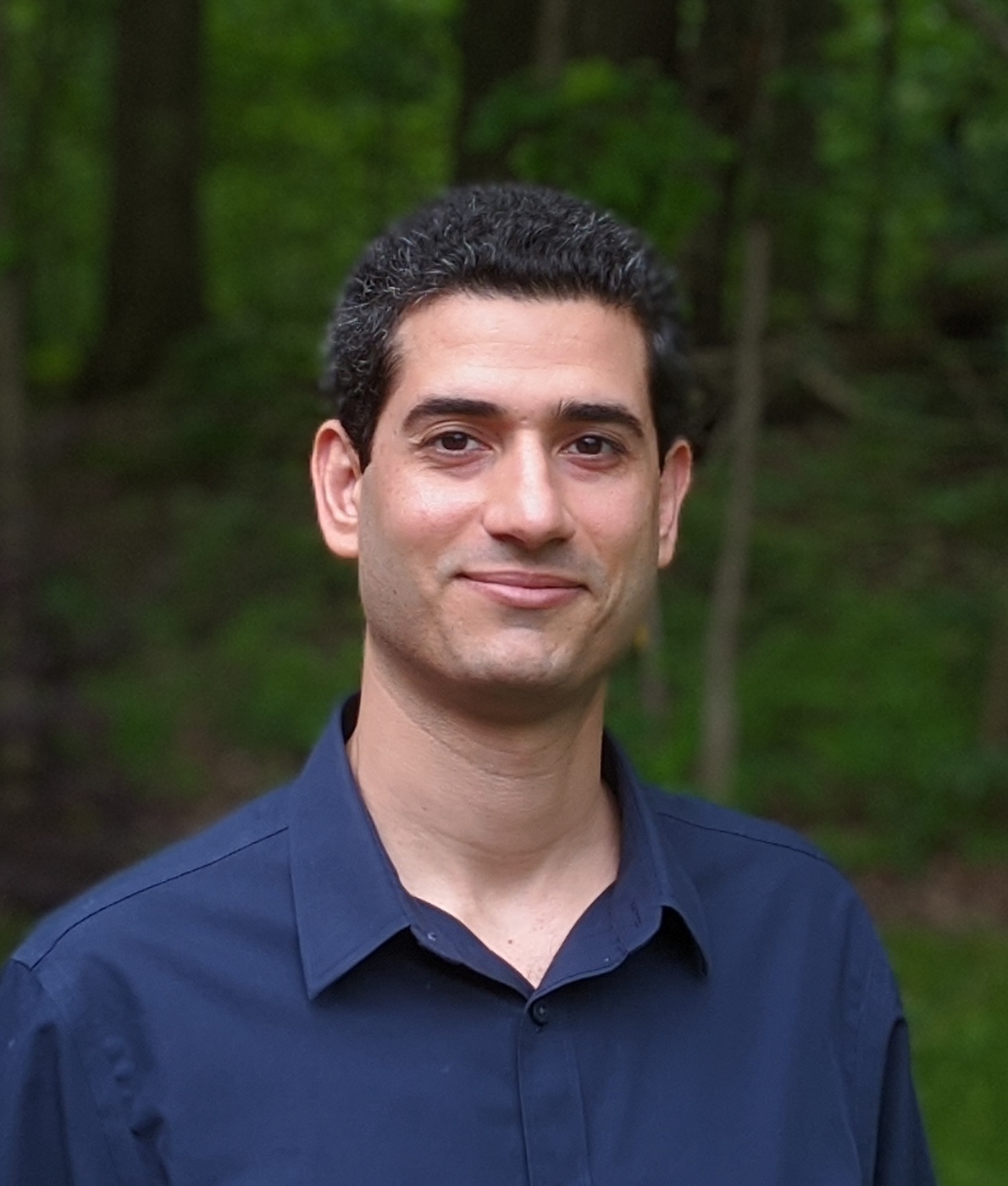}}] 
{Maani Ghaffari}
received the Ph.D. degree from the Centre for Autonomous Systems (CAS), University of Technology Sydney, NSW, Australia, in 2017. He is currently an Assistant Professor at the Department of Naval Architecture and Marine Engineering, and the Department of Robotics, University of Michigan, Ann Arbor, MI, USA. He recently established the Computational Autonomy and Robotics Laboratory. He is the recipient of the 2021 Amazon Research Awards. His research interests lie in the theory and applications of robotics and autonomous systems.
\end{IEEEbiography}

\begin{IEEEbiography}[{\includegraphics
[width=1in,height=1.25in, trim=0 0 0 0,clip,
keepaspectratio]{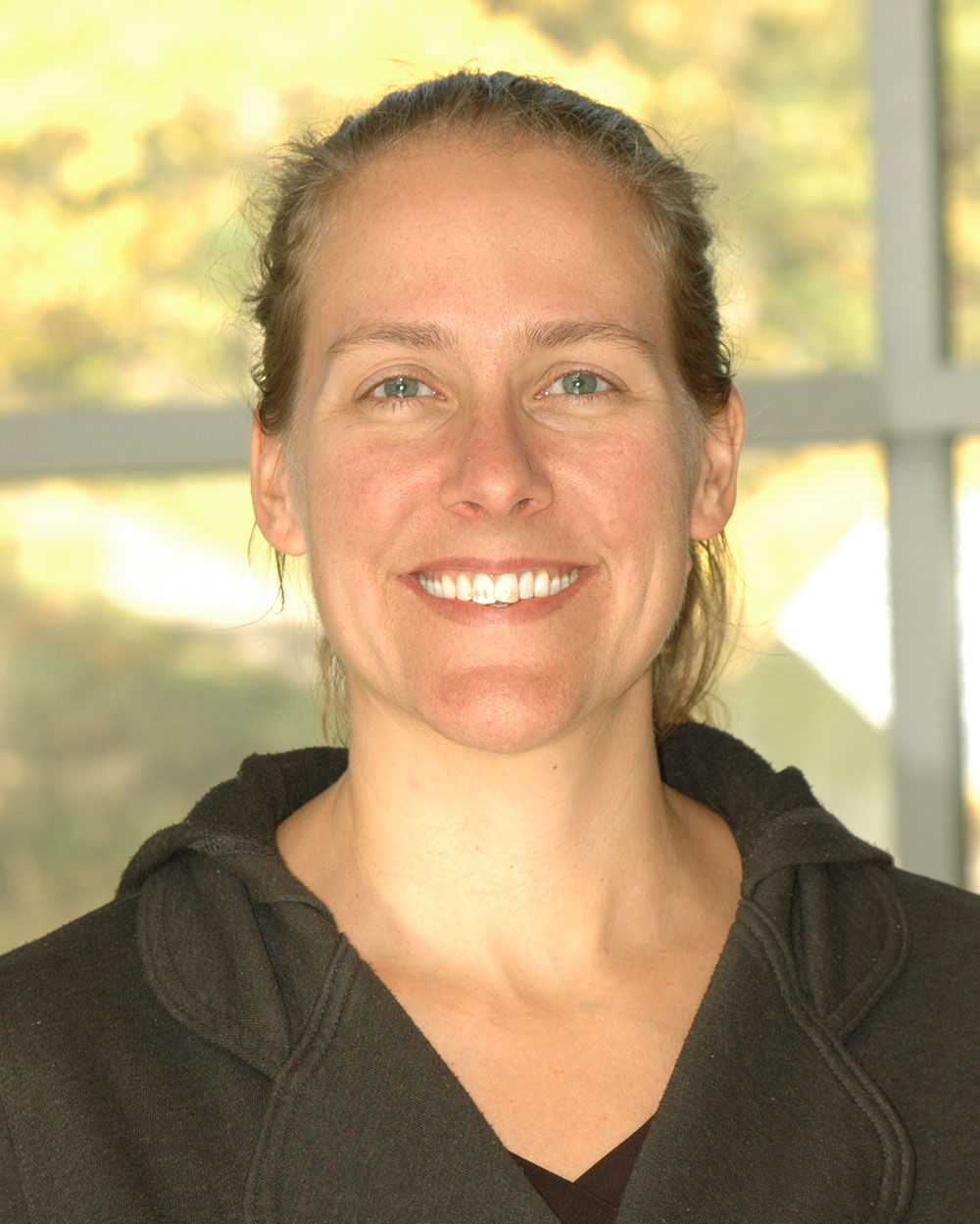}}] 
{Kira Barton}
is an Associate Professor in the Robotics Department and Mechanical Engineering Department at the University of Michigan. She received her B.Sc. in Mechanical Engineering from the University of Colorado at Boulder in 2001, and her M.Sc. and Ph.D. in Mechanical Engineering from the University of Illinois at Urbana-Champaign in 2006 and 2010. She is also serving as the Associate Director for the Automotive Research Center, a University-based U.S. Army Center of Excellence for modeling and simulation of military and civilian ground systems. She was a Miller Faculty Scholar for the University of Michigan from 2017 – 2020. Prof. Barton’s research specializes in advancements in modeling, sensing, and control for applications in smart manufacturing and robotics, with a specialization in learning and multi-agent systems.
\end{IEEEbiography}